\documentclass[letterpaper, 10 pt, conference]{ieeeconf}  

\IEEEoverridecommandlockouts                              

\usepackage{times}

\usepackage{multicol}
\usepackage[bookmarks=true]{hyperref}
\usepackage{graphicx,amsmath,algpseudocode,algorithm}
\usepackage{tabularx}
\usepackage[dvipsnames]{xcolor}
\usepackage{amsfonts}
\newtheorem{assumption}{Assumption}
\newtheorem{theorem}{Theorem}
\newtheorem{lemma}{Lemma}

\newcommand{\method}{\textsc{SOCRATES}}

\title{ 
SOCRATES: Text-based Human Search and Approach using a Robot Dog
}

\author{Jeongeun Park$^{1}$,  Jefferson Silveria$^{2}$, Matthew Pan$^{2}$ and Sungjoon Choi $^{1}$
\thanks{$^{1}$Jeongeun Park and Sungjoon Choi are with 
the Department of Artificial Intelligence, 
Korea University, Seoul, Korea
{\tt\footnotesize 
\{baro0906,sungjoon-choi\}@korea.ac.kr}.}%
\thanks{$^{2}$ Jefferson Silveria and Matthew Pan are with 
the Department of Electrical and Computer Engineering, Queens University, Kingston, Canada
{\tt\footnotesize \{jefferson.silveira,matthew.pan\}@queensu.ca}.}%
}

\begin{document}
\thispagestyle{empty}
\pagestyle{empty}
\maketitle

\begin{abstract}
In this paper, we propose a SOCratic model for Robots Approaching humans based on TExt System (\method) focusing on the \textit{human search and approach} based on free-form textual description; the robot first searches for the target user, then the robot proceeds to approach in a human-friendly manner. In particular, textual descriptions are composed of appearance (e.g., \textit{wearing white shirts with black hair}) and location clues (e.g., \textit{is a student who works with robots}). We initially present a Human Search Socratic Model that connects large pre-trained models in the language domain to solve the downstream task, which is searching for the target person based on textual descriptions. Then, we propose a hybrid learning-based framework for generating target-cordial robotic motion to approach a person, consisting of a learning-from-demonstration module and a knowledge distillation module. We validate the proposed searching module via simulation using a virtual mobile robot as well as through real-world experiments involving participants and the Boston Dynamics Spot robot. Furthermore, we analyze the properties of the proposed approaching framework with human participants based on the Robotic Social Attributes Scale (RoSAS) \cite{17_carpinella}. 
\end{abstract}

%
%
\section{Introduction}
In recent years, the coexistence of humans and robots has become increasingly common in daily life. Assistive robots are widely used in everyday commercial tasks within human environments such as delivering luggage at the airport or transporting packages such as food within an office. To ensure efficient human-robot interactions on such tasks, the ability for robots to interact with humans based on natural language and intuitive motions appears to be an important feature. In this paper, we propose a SOCratic model for Robots Approaching humans based on TExt System (\method) which focuses on the problem of \textit{human search and approach}. In this system, the target person is described via a free-form textual description; the robot first searches for a target person, then proceeds to approach in a human-friendly manner. 

Recent success on large pre-trained models, e.g., GPT~\cite{20_brown}, CLIP~\cite{21_radford}, has opened a new paradigm for robots to understand human languages, such as instruction following~\cite{22_ahn, 22_jang}, vision language-based navigation~\cite{20_zhu, 22_shah}, and zero-shot object search~\cite{22_gadre,22_khandelwal,22_majumdar,23_park}.
In this paper, we aim to build a robotic system that searches and approaches a target person based on a textual description of their appearance (e.g., \textit{person with black hair wearing a white shirt}) and location clues (e.g., \textit{is a student who works with robots}). Our task is similar to zero-shot object search~\cite{22_gadre,22_khandelwal,22_majumdar,23_park}, which aims to search for objects based on descriptions provided in free-form text, but differs in that our goal is to search for target \textit{people} rather than objects. Because our objective is to enable robots to search for people, we must also focus on correctly identifying them based on free-form user-entered text, as well as having a robot competently and appropriately interact with targets. 

Building on the success of large pre-trained models, Zeng et al.~\cite{22_zeng} proposed a modular framework called Socratic Model (SM). This framework leverages pre-trained models, which allows for an efficient and effective task-specific performance without fine-tuning on downstream task data. The SM framework formulates downstream tasks in a shared language domain between the pre-trained models and tackles the problem by building bridges among models from different modalities to facilitate communication. 
This approach has the advantage of leveraging pre-trained models trained on a large external database, which can avoid the need for task-specific data to train a single model. 
Inspired by the success of the SM framework, we propose to apply this approach to the task of \textit{human search and approach}. Our hypothesis is that the capabilities of large pre-trained models can be effectively leveraged to solve this task.

To this end, we propose a SOCratic model for Robots Approaching humans based on TExt System (\method), tackling the problem of \textit{human search and approach} where the target is given as a textual description. \method~consists of two serial modules: search and approach. The search module of \method, also referred to as a Human Search Socratic Model, connects large pre-trained models on a shared language domain to solve search downstream tasks without fine-tuning. In particular, it is composed of a Large Language Model (LLM), Vision and Language Model (VLM), and a waypoint generator. The LLM infers the searching prior based on the location clue of the target person, the VLM localizes the human with an appearance description, and the waypoint generator estimates the next waypoints for the robot to take based on the prior knowledge by LLM. 
The approach module is used after the search phase: the robot has to approach the target user in a comfortable and cordial manner, without being deemed threatening. To perform these functions, we propose a hybrid learning-based framework to estimate the cost function of the approach motion, composed of a learning from demonstration (LfD) step and knowledge distillation step from the large-language model.

The main contributions of the paper are as follows:
\begin{enumerate}
    \item We propose \method, which efficiently searches and approaches a target person where its description is given as \textit{free-form text}. 
    \item We present a Human Search Socratic Model, combining pre-trained models in the language domain to search for people based on textual descriptions.
    \item We propose a hybrid learning-based framework to generate a target-friendly approach motion composed of the learning from demonstrations (LfD) module and the knowledge distillation module from large language models. 
\end{enumerate}
%

%
%
\section{Related Work}

In this section, we provide existing work related to applying the SM to searching for objects and humans and approaching a person in a socially acceptable way.
Socratic Model (SM)~\cite{22_zeng} is a framework that enables composing pre-trained models to "talk to each other" so that the knowledge learned from a set of surrogate tasks can be applied to a new downstream target task. 
Zeng et al. \cite{22_zeng} proposed various applications of SM, including robotic perception and planning, open-ended reasoning on egocentric video, and multimodal assistive dialogue. SM has been widely used in robotics to solve language-based problems such as task and motion planning or visual navigation. 
Huang et al.~\cite{22_huang} proposed Inner Monologue for instruction-based task and motion planning, chaining Large-Language-Model (LLM) together with Vision-Language Models (VLM), robotic skills, affordance models, and human feedback in a shared language prompt. 
Shah et al.~\cite{22_shah} proposed LM-Nav for vision-language navigation tasks, which executes language instructions in a self-supervised manner, combining three different pre-trained models (i.e., LLM, VLM, and Vision-Language-Navigation). 


There have been recent attempts to search for an object described through free-form text, referred to as zero-shot object goal navigation \cite{22_gadre,22_khandelwal,22_majumdar,23_park}. Gadre et al.~\cite{22_gadre} has proposed a framework called 'CLIP on the Wheels', which leverages Gradient-weighted Class Activation Mapping (Grad-CAM)~\cite{16_selvaraju} of CLIP~\cite{21_radford} for localizing novel target objects and frontier based explorations~\cite{98_yamauchi} for exploration. 
Majumdar et al.~\cite{22_majumdar} proposed the method for this task, projecting navigation goals into a common semantic embedding space using CLIP~\cite{21_radford} to guide the agent to understand the image goal in language form. 
While some work has trained whole uni-model networks for text-based object search in simulated environments~\cite{22_majumdar,22_khandelwal}, their lack of application in real-world settings serves as a limitation.
Our task is similar to zero-shot object goal navigation as it is also based on text inputs; however, it differs in the fact that it requires adapting to the changing characteristics of the target person, as well as processing longer descriptions.

Approaching humans in a user-friendly manner is another key component in \textit{human search and approach}. Several papers~\cite{17_ahn, 21_konar, 20_chen} have covered the socially aware navigation problem with learning-based frameworks, as it is difficult to model complex social interactions explicitly. Ahn et al.~\cite{17_ahn} presented the method to comfortably approach a user by considering each user's personal comfort field, measured by existing theories in anthropology, and learning personalized comfort fields from users. 
Konar et al.~\cite{21_konar} proposed a sampling-based approximation to enable model-free inverse reinforcement learning for socially compliant navigations. 
In this paper, we contain a method based on a non-parametric inverse-reinforcement learning framework, this learning-based method has its strength in robustly learning the reward function with a small amount of noisy expert data.

%
%
\section{Human Search Socratic Model}

In this paper, we propose \method, which tackles the problem of text-based \textit{human search and approach}. The problem is separated into search and approach phases. Here, we present the Human Search Socratic Model, which connects pre-trained models from different modalities to shared language domains as a method to solve the downstream task of searching for a target person based on textual descriptions. 

\begin{figure*}[!t]
    \centering
    \includegraphics[width=0.95\textwidth]{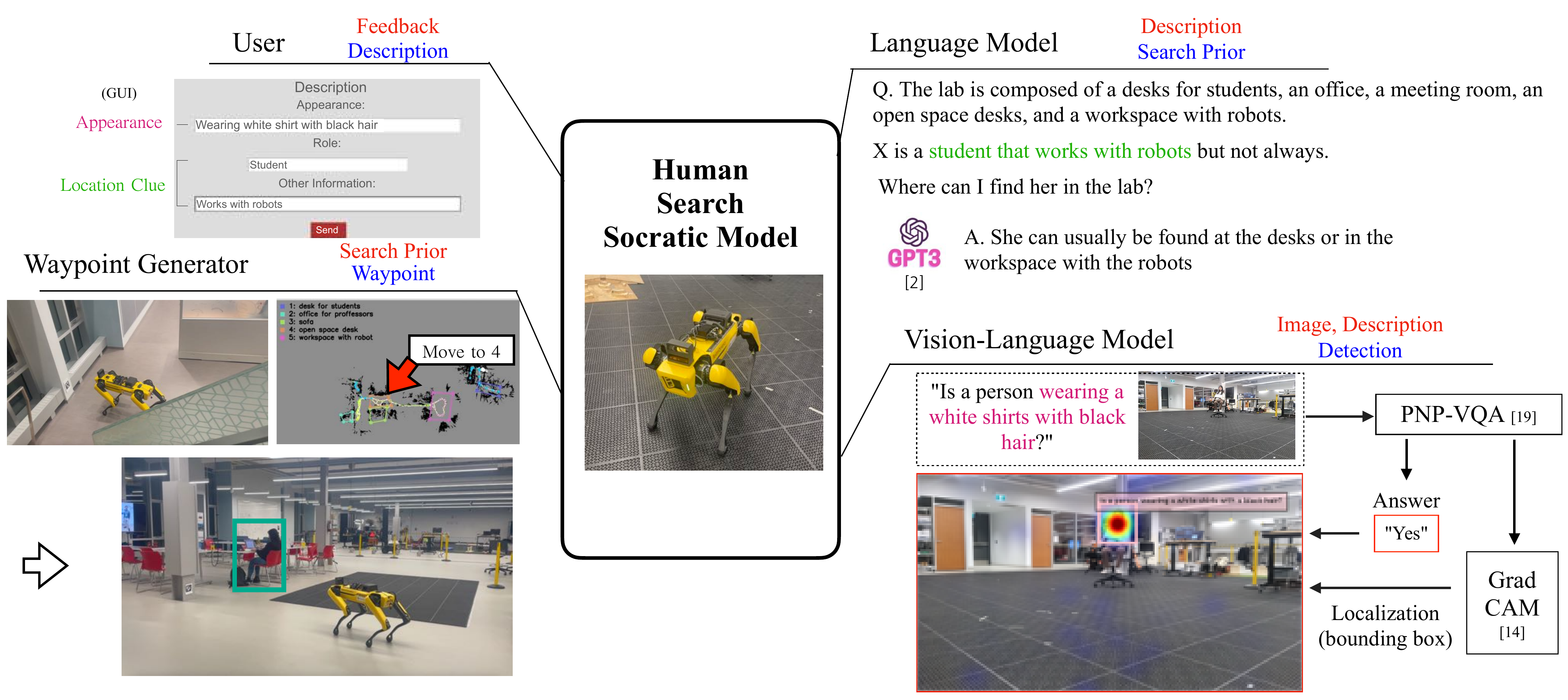}
    \caption{Overview of the proposed Human Search Socratic Model, \textcolor{red}{red}: input, \textcolor{blue}{blue}: output}
    \label{fig:search}
\end{figure*}

The Human Search Socratic Model is composed of three key components: a large-language model, a vision-language model, and a waypoint generator. The large-language model~\cite{20_brown} takes the description of a target person and estimates the search prior, whereas the vision-language model~\cite{22_tiong} processes images and descriptions to localize the target person. Lastly, the waypoint generator commands the robotic actions with the search prior and the detection results. An overview of the Socratic-based searching model is shown in Figure \ref{fig:search}. 

\begin{figure}[!t]
    \centering
    \includegraphics[width=0.47\textwidth]{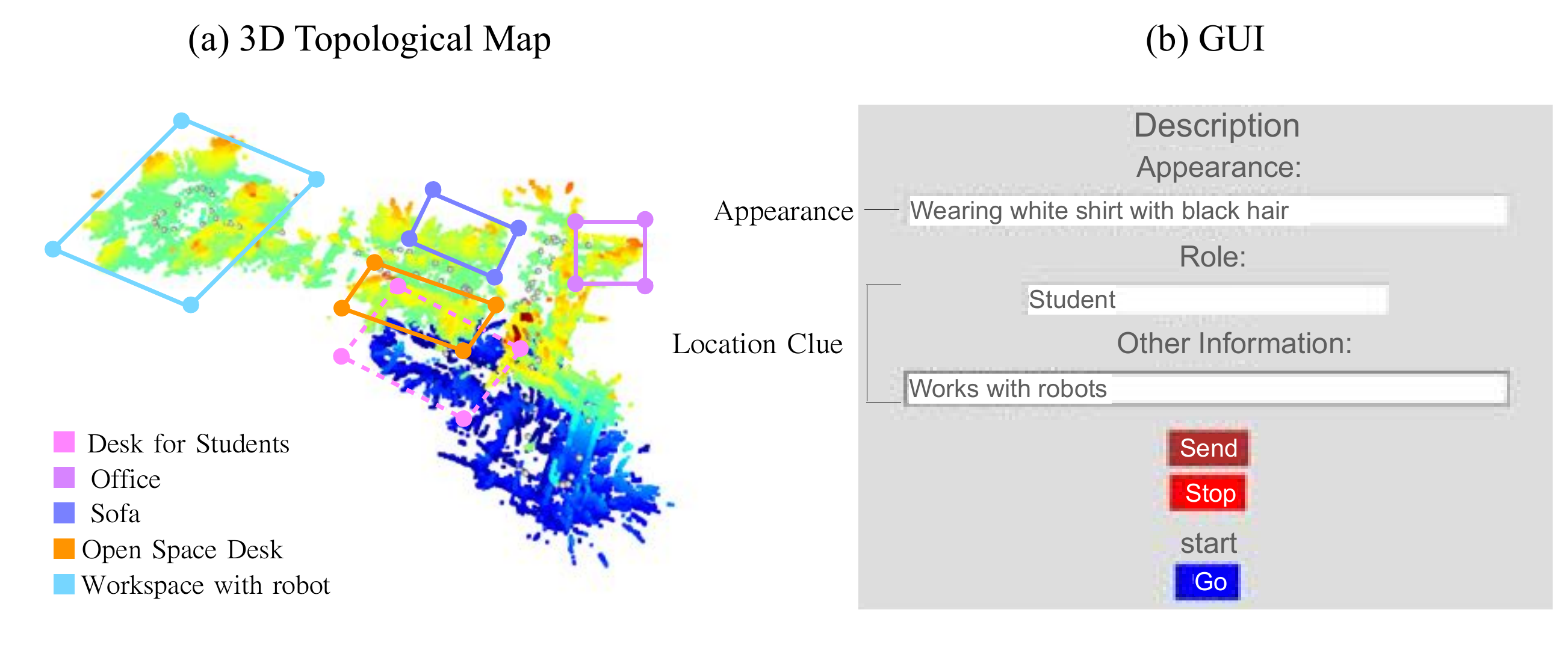}
    \caption{Input of the system.}
    \label{fig:input}
\end{figure}

%
%
\subsection{Problem Formulation}
In this framework, we tackle the problem of robotic human search based on textural descriptions $t$. Such descriptions are composed of appearance $t_1$ and location clues $t_2$ (e.g., \textit{wearing a white shirt with black hair} and \textit{is a student who works with robots}). We assume that an annotated map
and the location of the robot are accessible. The annotated map consists of area categories $\{a_i\}_{i=1}^K$ and the label of each area $a_i$. An example of the annotated map and the GUI used for textual description input of target people are illustrated in Figure \ref{fig:input}. The output of the Human Search Socratic Model is a waypoint $\mathbf{p} = (x,y,z,\theta)$, which directs the robot to a search location. 


\subsection{General Procedure}
Using large-language models, we estimate the search prior by computing the likelihood of the target person to occur in label $a_i$ of the annotated map shown in Figure \ref{fig:input}-(a). Based on this search prior, the waypoint generator calculates the cost of each label and generates the waypoint whose label has the lowest cost. Then, the waypoint generator conducts a local search of the area belonging to the chosen label using the detection results from the vision-language model. As the ambiguity between language and vision is inevitable, there is a possibility of obtaining multiple answers (potential target detections). 
As such, the system asks for user feedback on whether a detected person fitting the textual description provided is the target person. If not, the system continues the local search. When the waypoint generator finishes the local search, the system recalculates the cost of the map's label, moves to the next label, and starts another local search. The following subsections explain the details of each process. 

\subsection{Large Language Model} \label{sec:prior}
We leverage a large language model, i.e., GPT3~\cite{20_brown}, in order to provide a search clue to the robot. We gain commonsense knowledge of where the target person may be located on the annotated map. This knowledge guides the robot to search for the areas that have a high likelihood of the target person's placing. The prompt given to GPT3 is as follows: 
\begin{center}
\fbox{\begin{minipage}{8.5 cm} \textcolor{gray}{ \texttt{\small{
The lab is composed of [floorplan categories ($\{a_i, \cdots, a_K\}$)]. 
X is a [location clue ($t_2$)] but not always. 
Where can I find X in the lab?}}
}
\end{minipage}}
\end{center}
We denote a set of $M$ generated sentences $\{s_i\}_{i=1}^{M}$ as follows:
\begin{equation}
	\{s_i\}_{i=1}^M = f_1(t_2, a_1 , \cdots, a_K) \label{eq:prior}
\end{equation}
where $f_1$ is GPT3. 

\subsection{Vision-Language Model} \label{sec:vlm}
For the successful localization of the target person based on the description of appearance $t_1$, we utilize the vision-language model. In particular, we leverage PNP-VQA: a specific vision-question answering model proposed in ~\cite{22_tiong}. The input of the vision-language model is an image $I$ and a question, where the question is prompted as follows: 

\begin{center}
\fbox{\begin{minipage}{6.5 cm} \textcolor{gray}{ \texttt{ \small{
Is a person [appearance ($t_1$)]? }}
}
\end{minipage}}
\end{center}

As the vision-question-answering model provides an answer to a question, we process the answer (yes or no) as a binary label. Additionally, we leverage the Grad-CAM~\cite{16_selvaraju} of the VLM to localize the target person. Following the same approach in PNP-VQA~\cite{22_tiong}, we compute the relevance between the text and image features to obtain an activation map. 
Based on un-normalized Grad-CAM on the text and image similarity, we obtain the bounding box $\mathbf{b} = \{x_1,y_1,x_2,y_2\}$ of the image on the area where the Grad-CAM score is higher than the threshold.



\subsection{Waypoint Generator} \label{sec:search_wg}
The waypoint generator operates in two phases; the global search phase and the local search phase. The first phase determines which label of the annotated map to visit. The second module generates waypoints to search for the target person in the specific annotated area.

\paragraph{Global Search}
In the global search phase, the module estimates the likelihood of occurrence of the target person in the specific label based on the prior knowledge in Section \ref{sec:prior}. It then generates the waypoint to reach an area belonging to a specific label. 
The first step is to measure the occurrence score of label $a_i$ in the annotated map. We calculate the score with maximum word similarity~\cite{13_mikolov} between each word from the generated sentence in equation \ref{eq:prior} and the label name. The occurrence score becomes the average of each score calculated by generated sentences from $\{s_i\}_{i=1}^M$.

\begin{equation}
	p(a_i|t_2) = \frac{1}{M}\sum_{k=1}^M \max_l \ell(w(a_i), w(s_k^l))
\end{equation}
where  $s_k^l$ is a $l$-th word in sentence $s_k$, $\ell$ is a cosine similarity and $w$ is a word to vector embedding~\cite{13_mikolov}.
Based on the occurrence score, we compute the cost of visiting label $a_i$ for each label in the annotated map. The cost function is as follows: 

\begin{equation}
	c_l(a_i) = ||(\mathbf{p}_r  - \mathbf{p}_{a_i})||_2 + w_e (1- p(a_i|t_2))
        \label{eq:cost_search}
\end{equation}
where $\mathbf{p}_r$ is a current robot pose, $\mathbf{p}_{a_i}$ is the closest reachable position in the annotated area where its label is $a_i$, and $w_e$ is a weight of the occurrence score. Then the robot navigates to $\mathbf{p}_{a_i}$ that has the lowest cost among unvisited labels.

\paragraph{Local Search}
In the local search phase, the robot aims to search for a target person in the area that is annotated as the chosen label. After navigating to the area chosen as described in the global search phase, the robot conducts the local search. As detection based on language description is less reliable and also dependent on the specific viewpoint, we adopt an \textit{indirect search} approach. The robot searches for general humans first, obtains better viewpoints, and then detects the target based on the language description. We leverage the YOLOV5~\cite{2020_jocher} model pre-trained on the MS-COCO~\cite{14_lin} dataset to detect general humans. With an indirect search approach, we can prevent missing or false detection due to poor viewpoints and reduce the amount of feedback asked back to the user. 

The steps of the local indirect search are as follows. The robot first detects the general person, if any person is detected, the robot approaches or moves back to keep a 5m distance and centers the human to obtain a proper image. Then the vision-language model detects the person based on the description. If a potential target is detected, the system asks for user feedback. If no general person or text-based person is detected, the robot conducts frontier-based exploration~\cite{98_yamauchi} based on the map. We also use the history of the visited waypoints $\{\mathbf{p}_i\}_{i=1}^n$ to eject the waypoint that has already been visited. The robot visits the waypoint if the minimum $l2$-norm between past waypoints and the target waypoint is smaller than a certain threshold. With this indirect search approach, we were able to reduce the missing or false detection due to inappropriate viewpoints, hence reducing the amount of feedback asked back to the user. 


\section{Hybrid Learning-based Human Approach}

\begin{figure*}[!t]
    \centering
    \includegraphics[width=0.98\textwidth]{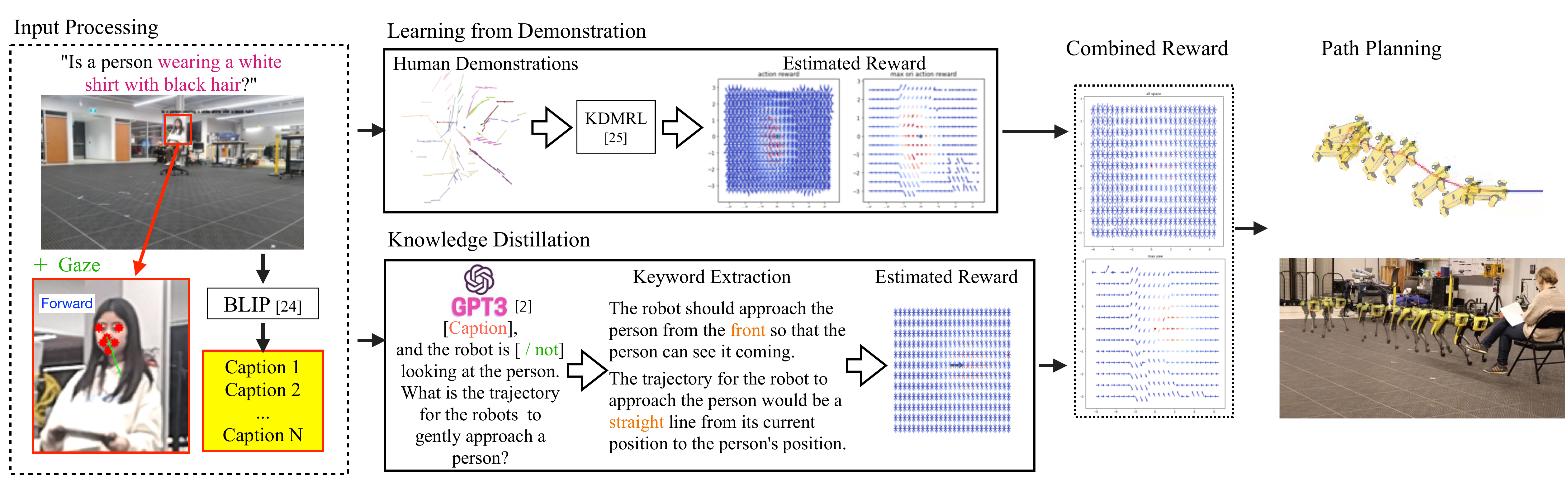}
    \caption{Overview of Proposed hybrid learning-based approach}
    \label{fig:approach}
\end{figure*}

In this section, we propose a hybrid learning-based framework for generating a cordial approach motion to a target person. Target-friendly approach motions are an essential part of the \textit{human search and approach} task since the robot has to move to a reachable position within the range of the target person without appearing threatening or distracting him/her. The proposed hybrid learning-based framework consists of two different modules: learning from demonstration (LfD) and knowledge distillation. Both modules estimate the reward function of the state space for a confiding approach trajectory and are combined to a final reward function. Next, the path planner conducts cost-aware planning based on the estimated reward function and sends actions (velocity commands) to the robot. The overview of the proposed hybrid learning-based framework is shown in Figure \ref{fig:approach}. 


%
%
\subsection{Problem Formulation} \label{sec:approach_pf}
The objective of the human approach is to generate a target-friendly approach motion. The inputs are estimated human pose $\mathbf{p}_h$, robot pose $\mathbf{p}_r$, captions $\{\mathbf{c}_i\}_{i=1}^N$, and gaze parameter $g$ which identifies whether the person is looking at the robot or not. The proposed framework first estimates the reward function $R$ of the state space, in which the frame of the reward function is defined as the robot pose frame with respect to the human pose frame. We define the state space as a combination of the two-dimensional robot poses relative to the human pose, gaze parameter, and the robot's speed; $\mathbf{x} = \{x_r^h, y_r^h, \theta_r^h, g, v\}$\footnote{We arranged as $x_r^h \in \{-6, -5.5, \cdots, 6\}$, $y_r^h \in \{-3, -2.5, \cdots, 3\}$  $\theta_r^h \in \{-\pi, -3/4\pi, \cdots, 3/4 \pi \}$ $v \in \{0.15, 0.4, 0.65\}$ with a total of $14976$ points and $7488$ inducing points.}. With the reward function $R$, the path planner plans the trajectory and controls the robot by velocity $(v_x,v_y, v_{\theta})$ commands. We assume the local egocentric map is accessible for cost-aware planning without collision.


\subsection{Input Processing}
The inputs of the human approach system are the pose of the robot relative to the pose of the person, gaze parameter, and image captions. As the detection result in section \ref{sec:vlm} can only estimate its position with depth information, we utilize a pre-trained human face mesh estimation network proposed in~\cite{19_lugaresi} to estimate the orientation of a target person. The mesh detection is conducted on a cropped image of the target's head based on the detection result. If the absolute orientation of the human face is less than 40 degrees, the gaze parameter becomes $1$; otherwise, $0$. Furthermore, to process the image in the format of the input of the language model, i.e., text, we estimate the caption by the image-captioning branch (BLIP \cite{22_li}) of PNP-VQA~\cite{22_tiong}.  

\subsection{Learning from Demonstration} \label{sec:lfd}
As it is hard to explicitly model complex social behaviour for the robot to create appropriate and non-threatening approaching motions, we adopt a learning from demonstration (LfD) paradigm. In particular, we leverage the Kernel Density Matching Reward Learning~\cite{16_choi} (KDMRL) framework to estimate the reward function of the approaching motion. It optimizes the reward function that matches the state-action distribution of expert demonstrations.

In the density matching reward estimation framework, we assume that the demonstrations are sampled from stationary state distribution $\mu(\mathbf{x})$. 
Following the objective of the inverse reinforcement learning formulation, the objective of density matching reward learning (DMRL) is to maximize the dot product of the density of the state space $\hat{\mu}$ and the reward $R$ subject to $||R||_2 \leq 1$. Then, the problem can be relaxed by solving the following optimization problem:
\begin{equation}
	\underset{R}{\text{maximize}} \  \tilde{V}(R) = \sum_{\forall \mathbf{x} \in U} \hat{\mu}(\mathbf{x})\tilde{R}(\mathbf{x}) - \frac{\lambda}{2} ||\tilde{R} ||^2_{\mathcal{H}}
	\label{eq:kdmrl_re}
\end{equation}
where $\tilde{R}$ is a optimization subject, $\mathbf{x}$ is a state, $\lambda$ is a smoothness parameter, $U = \{\mathbf{x}_i^U\}_{i=1}^{N_U}$ is a set of $N_U$ inducing points, and $||\tilde{R} ||^2_{\mathcal{H}}$ is the squared Hilbert norm. Then the reward function with inducing points is as follows:
\begin{equation}
	\tilde{R} (\mathbf{x}) = \sum_{i=1}^{N_U} \alpha_i k(\mathbf{x},\mathbf{x}_i^U) \label{eq:kdmrl_re2}
\end{equation}
where $\mathbf{x} \in \mathcal{X}$, $\mathcal{X}$ is the state space, $\mathcal{U} \subset \mathcal{X}$ is a set of pre-defined inducing points, $\alpha \in \mathbb{R}^{N_U}$ is a parameter determining the shape of the reward function, and $k(\cdot, \cdot)$ is a positive semi-definite kernel function.
Kernel density estimation is used to estimate the probability density function of $\mu$, 

\begin{equation}
	\hat{\mu} (\mathbf{x}) = \frac{1}{Z} \sum_{k=1}^{N_D} \cos(\frac{\pi}{2} (1-\gamma_k)) k_{\mu}(\mathbf{x},\mathbf{x}_k^D) \label{eq:kdmrl_mu}
\end{equation}
where $N_D$ is the number of training samples and $\mathbf{x}_k^D$ is the $k$-th training data, $\gamma_k$ is a leverage value, and $k_{\mu}(\cdot, \mathbf{x}^D)$ is a kernel function whose integral is one. The leverage value $\gamma_k = \delta^{T-t}$ indicates the weight of $k$-th data with $\delta \in [0,1]$, $T$ and $t$ is the length and time-step of the trajectory of the data and $Z$ is a normalization constant. 

With equation \ref{eq:kdmrl_mu}, \ref{eq:kdmrl_re2}, and norm regularization $\frac{\beta}{2} ||\alpha||_2^2$, we can formulate equation \ref{eq:kdmrl_re} as matrix form as follows
\begin{equation}
	\max_a \tilde{V} =\frac{1}{Z} \alpha^T K_U K_D \mathbf{g} - \frac{\lambda}{2} \alpha^TK_U\alpha - \frac{\beta}{2}\alpha^T\alpha 
	\label{eq:kdmrl_for}
\end{equation}
where $[K_U]_{ij} = k(\mathbf{x}_i^U, \mathbf{x}_j^U)$, $[K_D]_{ij} = k_{\mu}(\mathbf{x}_i^U, \mathbf{x}_j^D)$, and $\mathbf{g} = \{\cos(\frac{\pi}{2} (1-\gamma_k))\}_{k=1}^{N_D}$ is a vector of leverage values. The analytic solution of equation \ref{eq:kdmrl_for} with respect to $\alpha$ is as follows:

\begin{equation}
	\hat{\alpha} = \frac{1}{Z} (\beta K_U + \lambda I)^{-1}K_U K_D \mathbf{g}
\end{equation}


\subsection{Knowledge Distillation}\label{sec:kd}
Furthermore, we distillate the knowledge from the Large-Language model \cite{20_brown} to generate the approaching motion of the robot. We extract commonsense knowledge from GPT3 \cite{20_brown} to estimate the reward function. 
With gaze parameter $g$ and captions $\{\mathbf{c_i}\}_{c=1}^N$, we make prompt for GPT3 as follows:

\begin{center}
\fbox{\begin{minipage}{8.5 cm} \textcolor{gray}{ \texttt{ \small{
[caption ($\mathbf{c_i}$)] and the robot is [([$g=1$)/ not ($g=0$)] looking at a person. What is the trajectory for the robots to gently approach a person?
}}
}
\end{minipage}}
\end{center}

We generate sentences with GPT3, $\{s^t_i\}_{i=1}^{n\cdot N}$, $n$ times for each $N$ captions. Then, we extract the keywords\footnote{\texttt{straight, 45, side, behind, curve, front, curved} for the position, \texttt{slow, slowly} for velocity, and \texttt{not} for negative sign.} that are relevant to the trajectory, i.e., words related to the position, velocity, and negative sign. The keywords extraction function is denoted as $f_k(s^t_i)$ and extracted keyword as $\{w^l_i\}_{l=1}^L=f_k(s^t_i)$. 

To generate a trajectory based on a set of keywords, we set a word-to-trajectory function $f_d(w)$ for generating a partial path based on the word. The detail of the function is explained in the supplementary materials. Based on this function $f_d(w)$, we generate an estimated trajectory by iterating over the extracted keyword and stacking the partial path from the word. Next, we transfer the generated trajectories to the reward function; the procedure of estimating the reward function is shown in algorithm \ref{alg:cm_reward}. Following reward estimation, we smooth the reward function with the radial basis kernel function $k_{r}(\cdot, \cdot)$. The knowledge-distillation-based reward function becomes $R = k_r(\mathbf{x}, \mathbf{x}) \cdot R$.

\begin{algorithm}
\caption{Reward Estimation for Knowledge Distillation Based Human Approach}\label{alg:cm_reward}
\begin{algorithmic}
\State $R = [0]_{s}$ where $s$ is a dimension of the state space
\For {i = 0 to $n\cdot N$}
    \State  $\{w^l_i\}_{l=1}^L=f_k(s^t_i)$, $x = 0$, $y=0$
    \For {$l$ = $L$ to $0$}
        \State $\{d\mathbf{x}^j, d\mathbf{y}^j\}_{j=1}^{N_L} = f_d(w^l_i)$ // extract keyword
        \State $v=0.15$ if \{$\text{'slow', 'slowly'} \in \{w^l_i\}_{l=1}^L$\} otherwise $v=0.6$ 
         // set velocity
        
        \For {j = 0 to $N_L$}
            \State $\theta = \arctan(d\mathbf{y}^j[-1],d\mathbf{x}^j[-1]) $ \
            \State $R(x+d\mathbf{x}^j, y+d\mathbf{y}^j,[\theta]_{d},\cdot,[v]_{d}) += 1$ \
            \State // update reward function \
            \State $x \leftarrow x + d\mathbf{x}^j[-1]$, 
                $y \leftarrow y + d\mathbf{y}^j[-1]$ 
            \State // update last position \ 
        \EndFor
    \EndFor
\EndFor
\State $R \gets R/(N\cdot n)$
\end{algorithmic}
\end{algorithm}

\subsection{Path Planner}
In the path planning step, the main goal is to generate paths based on the reward function for any initial configuration of the robot.
To achieve this, we apply the FMT* path planning algorithm \cite{janson2015fast} made available in the Open Motion Planning Library (OMPL) \cite{sucan2012ompl} in an environment where the cost of each state is a function of the rewards obtained through the hybrid learning-based reward estimation.

With reward functions from the learning-based framework $R_I$ and knowledge distillation-based reward function $R_L$, we estimate the combined reward function by weighted summation of the two parts, $R_T = w_r R_I + (1-w_r) R_L$ such that $R_T \in [0,1]$. In this system, we set $w_r=0.2$. 
The resulting function is converted into a cost $1- R_T$ that is used by FMT* to penalize undesirable states. 
We set the cost function to take into account the reward $R_T$ and the path length of the arriving state. The cost function is as follows:
\begin{equation}
    c_m(\mathbf{\bar{x}}_1,\mathbf{\bar{x}}_2) = \zeta (1- R_T(\mathbf{\bar{x}}_2))\textrm{dist}(\mathbf{\bar{x}}_1,\mathbf{\bar{x}}_2),
\label{eq:approach_cost1}
\end{equation}
\begin{equation}
    \textrm{dist}(\mathbf{\bar{x}}_1,\mathbf{\bar{x}}_2) = w_p||\mathbf{q}_2-\mathbf{q}_1|| + w_o|\theta_2 - \theta_1|,
\label{eq:approach_cost2}
\end{equation}
where $\mathbf{\bar{x}}_i = \{\mathbf{q}, \theta\}$ is the $i$-th configuration of the robot such that $\mathbf{q} \in \mathbb{R}^2$ and $\theta \in SO(2)$. The weights $w_p$ and $w_o$ are used in the distance function to give more or less importance to each component of the state. $\zeta$ is a parameter that is used to fine-tune whether the solution converges towards the shortest path or to one that closely matches the reward function according to its magnitude. 

\section{Experiments}
In this section, we analyze the properties of the proposed method and discuss the experimental results. We validate our search module in the real world with human participants while also conducting an additional experiment on simulated environments. In addition, we conduct human-subject pilot studies on the proposed approaching motion of our system. The details of hyperparameters and the environments are explained in the supplementary materials.

\begin{figure}
    \centering
    \includegraphics[width=0.95\linewidth]{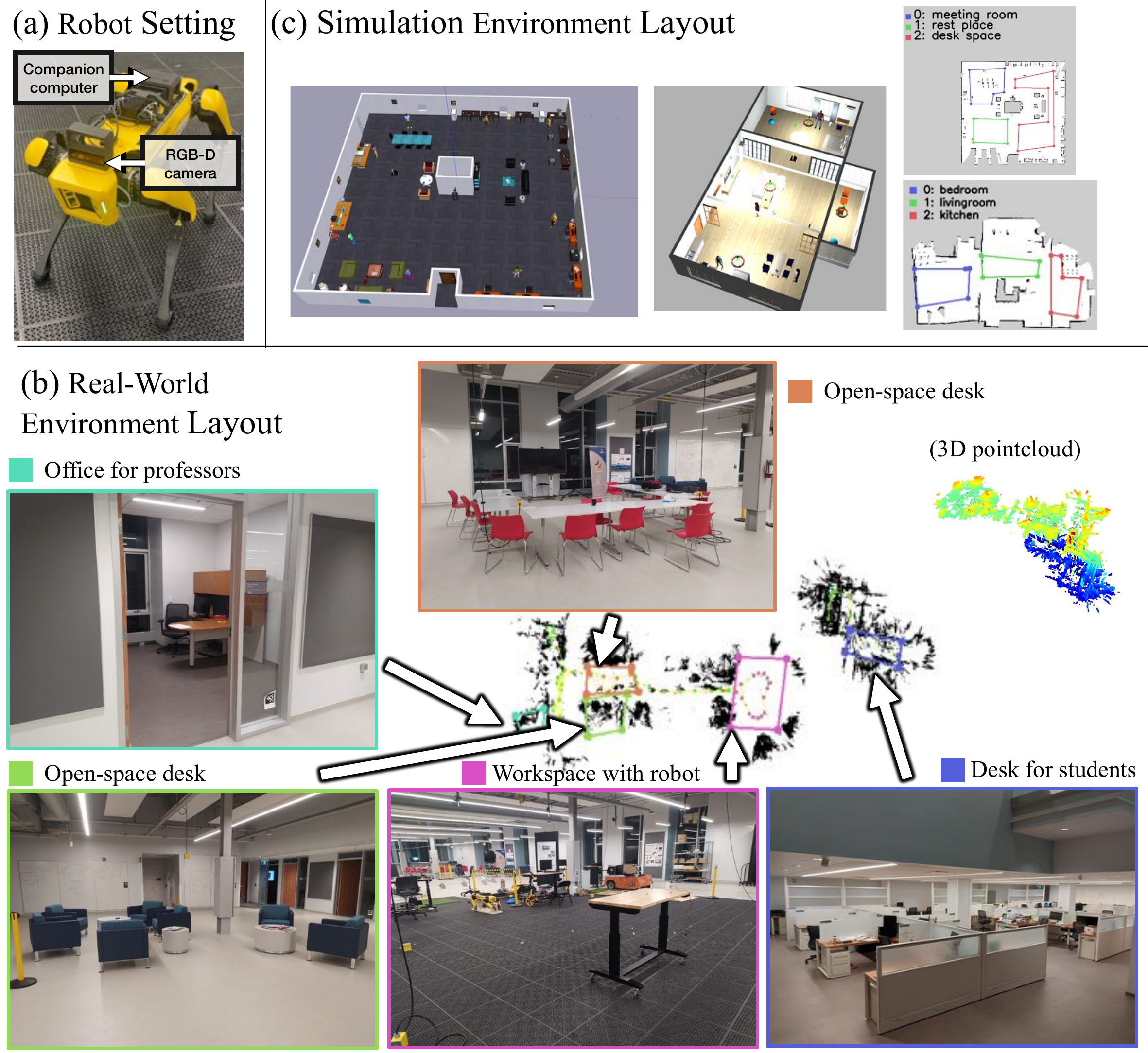}
    \caption{Robot settings and Environmental settings for Search Evaluation.}
    \label{fig:setup}
\end{figure}

\begin{figure*}[!t]
    \centering
    \includegraphics[width=0.9\textwidth]{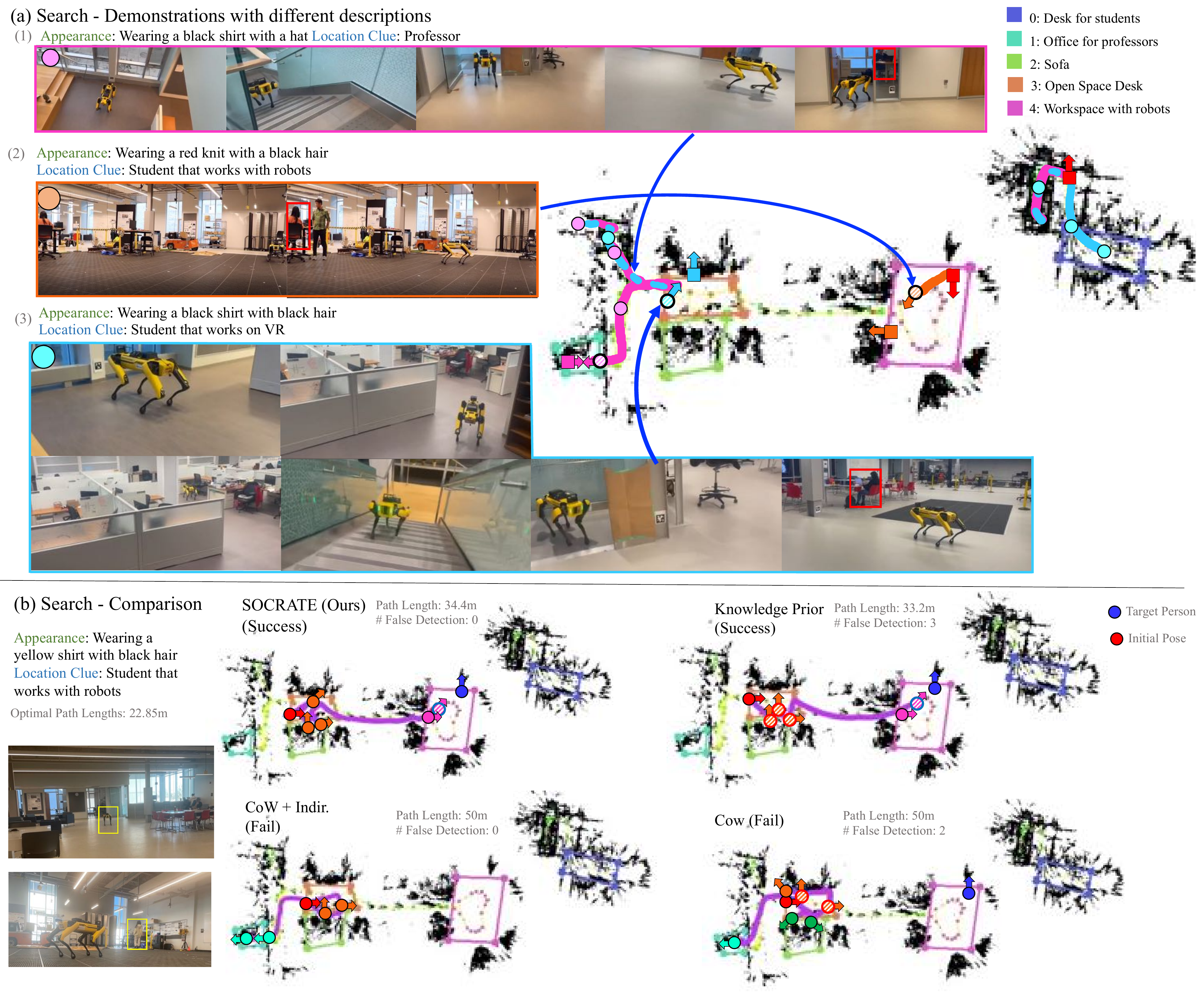}
    \caption{Illustration of the Experimental Results. (a) shows the exemplary demonstrations of the proposed search module. The circles represent waypoints. (b) shows how the trajectories vary between methods. Circles represent waypoints and the circles with hashes represent the viewpoint that contained the target detection, where the red line denotes false detection and the blue line denotes true detection.}
    \label{fig:exp}
\end{figure*}

%
%

\begin{table}[!t]
    \centering
    \begin{tabular}{c|c c c c}
         \hline
         Method  & SPF $\uparrow$& SPL $\uparrow$ & SR $\uparrow$ & \# FD $\downarrow$\\
         \hline
         CoW \cite{22_gadre} & 0.214 & 0.314 & 0.4 & 0.8 \\
         CoW \cite{22_gadre} $+$ Indir.  & 0.247 & 0.295 & 0.6 & 0.3 \\
         Knowledge Prior \cite{19_zhang} & 0.55 & 0.636 & 0.9 & 0.8 \\
         \method (Ours) & \textbf{0.597} & \textbf{0.647} & \textbf{1.0} & \textbf{0.2} \\
         \hline
    \end{tabular}
    \caption{Result of Search Module on Real-world Environment.}
    \label{tab:seaech_r}
\end{table}

\subsection{Real-world Search Experiments} \label{sec:rwse}

In this section, we will discuss the experimental results of the proposed Human Search Socratic Model in the real-world environment. We use a Spot robot mounted with a Luxonis OAK-D-Pro camera~\cite{21_oakd} as shown in Figure \ref{fig:setup}-(a). The real-world environment setting is a robotics lab environment shown in Figure \ref{fig:setup}-(b), Five different annotations of the floorplan are used to distinguish locations in the lab. We conducted our real-world experiments using ten different locations, with five different people and two unique initial positions for each person. The performance of the system is evaluated using three metrics: success rate (SR), success rate per path length (SPL), and the number of false detections (\# FD). SR and SPL are widely used in the evaluation of visual search tasks, as reported in previous studies such as \cite{18_anderson}. The \# FD metric is the average number of false detections, equivalent to the number of false feedback from the users. Success is defined as the robot successfully localizing the target person within a 5 m radius and receiving positive feedback from the users. A trial is marked as a failure if the system does not meet the success criteria, in total path length less than 30 m and \# FD less than five.

We compare the proposed system with three different approaches: \textit{CoW}, \textit{CoW + Indir}, and \textit{Knowledge prior}. CoW~\cite{22_gadre} leverage direct search without prior knowledge, adapting only the distance term in equation \ref{eq:cost_search}, and replacing the Grad-CAM~\cite{16_selvaraju} of CLIP~\cite{21_radford} with BLIP~\cite{22_li}. In addition, we compared the proposed method with CoW~\cite{22_gadre} with indirect search (CoW + Indir.). \textit{Knowledge Prior} is the method that solely utilizes the vision-language model without general 'human' detection, which resembles previous work \cite{19_zhang,09_kollar}. If the target is detected, it approaches within 5m of the target and requests user feedback; otherwise, the robot explores based on FBE~\cite{98_yamauchi}


The performance of the proposed search module is reported in Table \ref{tab:seaech_r}. The system proposed in this work has the best performance on the SPL metric with a gap of $0.137$. While the direct search has the highest success rate, the gap of $0.1$ is not significant. 
In addition, the direct search has more false detections compared to the indirect search, with a gap of $0.6$. As the proposed method obtains a human-centered and well-scaled image, this property leads to a decrease in the number of false detection and an increase in SPL. Compared to methodologies without prior knowledge for searching, methods that utilize prior knowledge have a significantly higher SPL metric, with an average $0.23$ gap. As such, we would like to posit that prior knowledge plays an essential role in the efficient search by robots. 

\begin{figure}[]
    \centering
    \includegraphics[width=0.95\columnwidth]{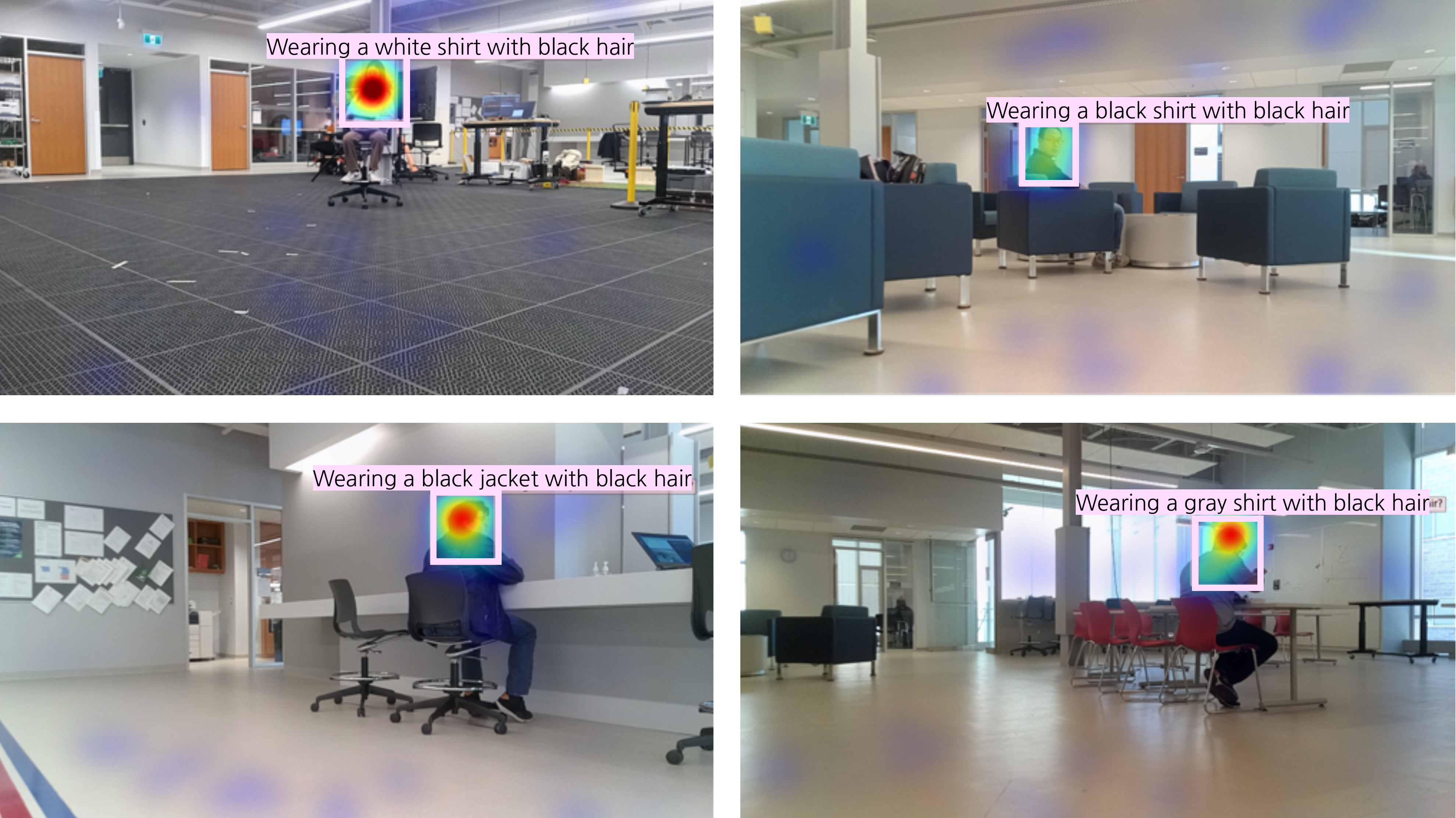}
    \caption{Detection Results}
    \label{fig:det}
\end{figure}

Figure \ref{fig:exp}-(a) and (b) shows the trajectory of the robot. Figure~\ref{fig:exp}-(a) illustrates three different demonstrations of the proposed searching method. The robot estimates the area where the target person is likely to be located first and then performs a local search within the likely area. For example, in Figure \ref{fig:exp}-(a)-(2), the robot infers, according to Section \ref{sec:prior} that the target person (a student that works on VR headsets) is likely to be located in the lower-floor student cubicle area or upper-floor open-space where the VR headsets are stored. The robot searches for the target within the lower-floor cubicles first, as it is closer to its current position, and then moves to the upper-floor space when the target is not detected.
Figure~\ref{fig:exp}-(b) shows the trajectory with different methods. With prior knowledge, the robot searches the upper-floor open-space first and then moves to the workspace with robots. On the other hand, without prior knowledge, the robot searches in the order of open-space area, sofa, and offices, leading to failure. Although the proposed indirect search does not show a significant gap in path length with direct search, the indirect search asks for less feedback from the user. Furthermore, Figure \ref{fig:det} shows the example of the system detecting the target person based on their appearance. 

\subsection{Simulation Search Experiments}
To show the expansion of the proposed method to various scenes other than the lab, we evaluate the system in two more simulation environments, i.e., a house and an office. We utilized a virtual Pioneer3AT robot mounted with an RGB-D camera and Lidar within the Gazebo simulator. We ran 24 experiments, two different initial positions for each person, with four different people in the house setting and eight in the office setting. We assume all people are static in the simulation environment. The comparison methods and metrics used are the same as Section \ref{sec:rwse}.
When the total path length is more than $15$ or the number of false detection is larger than three, we deem the trial as a failure.

 %
 %

\begin{table}[!t]
    \centering
    \begin{tabular}{|c|c |c |c| c|}
         \hline
         Method & SPF $\uparrow$& SPL $\uparrow$ & SR $\uparrow$ & \# FD $\downarrow$\\
         \hline
         CoW \cite{22_gadre} & 0.304 &0.379 & 0.423 & 0.69 \\
         CoW \cite{22_gadre} $+$ Indir. & 0.352 & 0.50 & 0.625 & 0.66 \\
         Knowledge Prior \cite{19_zhang} & 0.460 & 0.615 & 0.833 & 0.375 \\
         \method (Ours) & \textbf{0.564} &  \textbf{0.709} & \textbf{0.956} & \textbf{0.260} \\
         \hline
    \end{tabular}
    \caption{Result of Search Module on Simulation Environment.}
    \label{tab:search_s}
\end{table}

 The proposed method outperforms the compared method in all of the metrics. The number of false detection is higher with a $0.08$ gap compared to the direct search method. In addition, the result shows $0.041$ higher SPL on indirect search compared to direct search. The method with prior knowledge increases the performance, $0.32$ in SPL and $0.178$ in SR. The success rate of indirect search is higher than direct search on the simulational result, but the gap of $0.041$ is minor. Unlike real-world experiments, the number of false detection increases when we do not adapt the prior knowledge. Since the simulation environment is more crowded and contains more overlapping clothes, the longer path length leads to more false detections.

\subsection{Real-world Approach Experiments}

\begin{figure}[!t]
    \centering
    \includegraphics[width=0.5\textwidth]{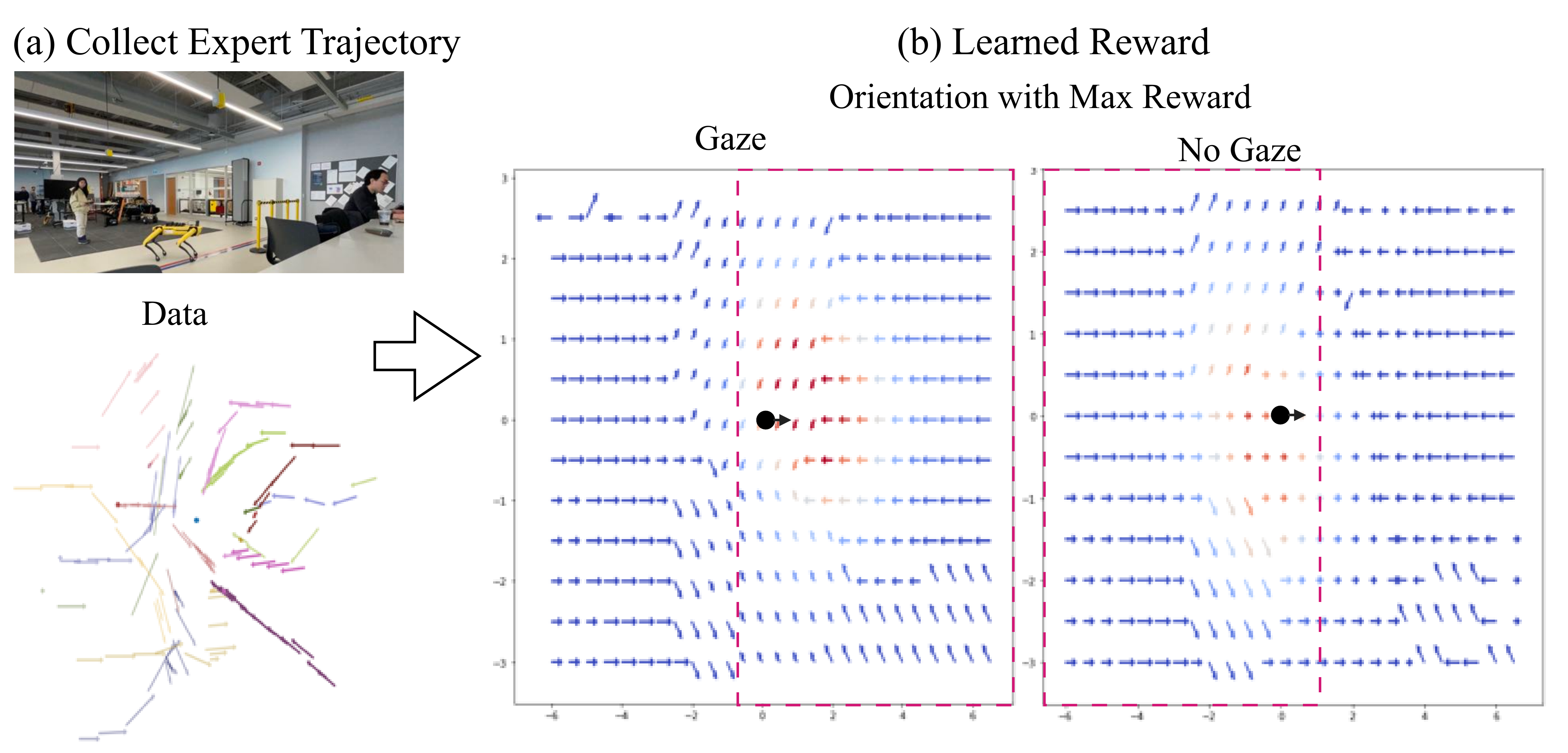}
    \caption{The illustration of data collection for the Learning-from-Demonstration module and the resulting reward function. The arrow heading indicates the orientation of the robot and the size indicates the speed. We only plot the reward that has a maximum reward on corresponding orientation and speed.}
    \label{fig:irl}
\end{figure}

\begin{figure*}[!t]
    \centering
    \includegraphics[width=0.9\textwidth]{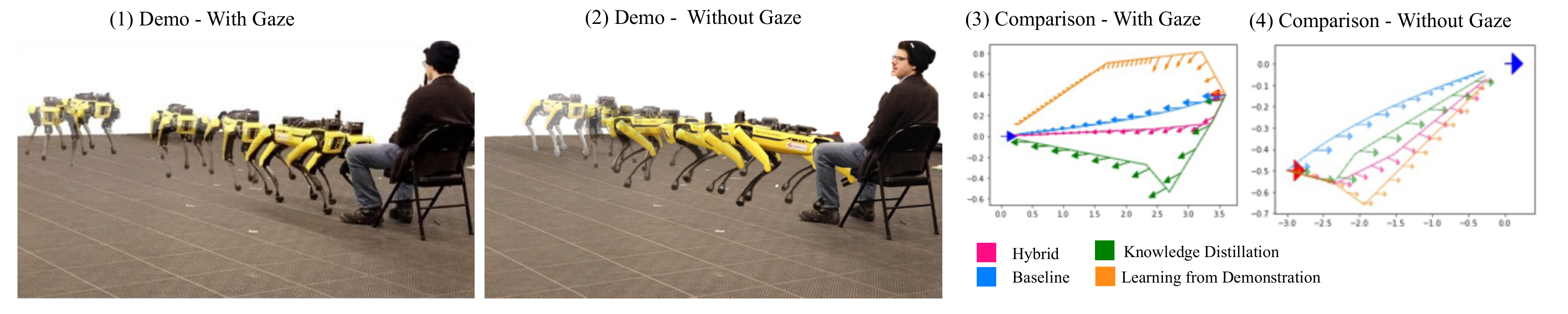}
    \caption{Illustration of the Experimental Results on approach phase.}
    \label{fig:exp_app}
\end{figure*}

\begin{table}[t]
    \centering
    \begin{tabular}{|c|c|c|c|}
        \hline
        Method & Competence $\uparrow$ & Warmth  $\uparrow$ & Discomfort $\downarrow$ \\
        \hline
         Baseline & 4.41 (1.08) & 3.34 (1.23)& 3.23 (1.23)\\
         KD & 3.96 (0.82) & 3.28 (1.03) & 3.59 (1.08)\\
         LfD & 4.14 (1.01) & 3.28 (1.29) & 3.30 (1.54) \\
         Ours & \textbf{4.56} (1.28) & \textbf{3.63} (1.28) &\textbf{2.96} (1.28)\\
         \hline
    \end{tabular}
    \caption{RoSAS Scale results in the total pool}
    \label{tab:rosas_1}
\end{table}

\begin{table}[t]
    \centering
    \resizebox{0.85\columnwidth}{!}{%
    \begin{tabular}{|c|c|c|c|c|c|c|c|}
        \hline
         Method&
        \multicolumn{2}{|c|}{Competence $\uparrow$} & \multicolumn{2}{|c|}{Warmth $\uparrow$} & \multicolumn{2}{|c|}{Discomfort $\downarrow$} \\
        &  Exp & IExp & Exp & IExp & Exp & IExp \\
        \hline
         Baseline & 4.30 & \textbf{4.58} & 3.11 & \textbf{3.58} & 3.35 & 3.12\\
          KD & 3.94 & 4.14 & 3.17 & 3.02 & 3.41 & 3.36 \\
          LfD & 4.24 & 3.99 & 3.63 & 2.94 & 3.25 & 3.36 \\
          Ours & \textbf{4.71} & 4.41 & \textbf{3.80} & 3.46 & \textbf{2.98} & \textbf{2.90}\\
         \hline 
    \end{tabular}%
    }
    \caption{RoSAS Scale results depending on the group.}
    \label{tab:rosas_2}
\end{table}

\begin{figure*}[!t]
    \centering
    \includegraphics[width=0.99\textwidth]{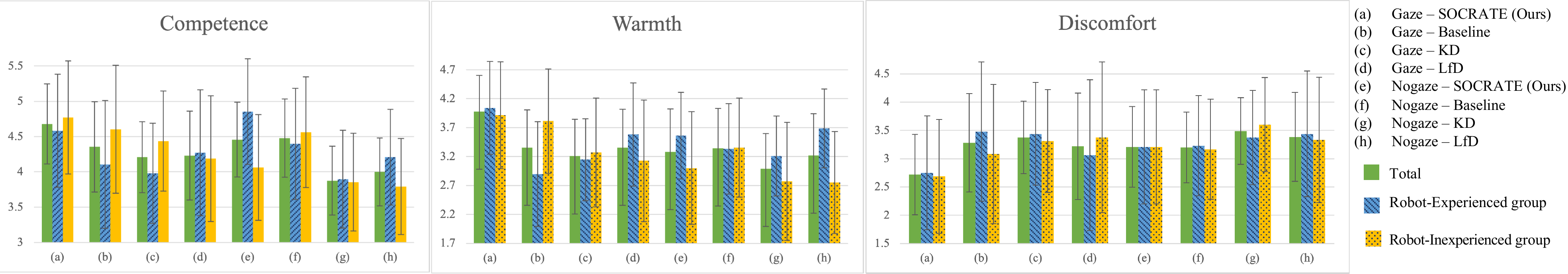}
    \caption{Average rate of the RoSAS attributes from different groups with 95\% confidence bound.}
    \label{fig:rosas}
\end{figure*}

In this section, we conduct a real-world human subject pilot study to analyze the properties of the proposed approach module with the SPOT robot in Figure \ref{fig:setup}-(a). The robot approaches the person from 5 m away to closer than 0.6 m, which is inside the reachable space for the target person. We evaluate the proposed method based on the Robotic Social Attribute Scale (RoSAS) \cite{17_carpinella}. We compare the proposed approaching motion with three different methods of approach generation: learning from demonstration (LfD), knowledge distillation (KD), and a baseline condition (having the robot directly move toward the target person based on position control). The LfD-based approach only leverages the reward function described in Section~\ref{sec:lfd}, and the KD-based approach solely adapts the reward function shown in Section~\ref{sec:kd}. 

We recruited 16 people (gender: M=10, F=6, ages: 21-47) to participate in the approach experiments.  We divided the participants into two groups: an experienced group containing those who have worked with robots previously (n=8) and an inexperienced group (n=8). The participants were asked to rate the robot's approach motion using 18 word descriptors as part of the RoSAS inventory, which measures user perceptions of the robot's behaviour along the dimensions of \textit{competence}, \textit{warmth}, and \textit{discomfort}. Each participant reported how applicable each keyword described their perception of the robot under different robot conditions on 7-point Likert scale that ranged from 1 (not at all) to 7 (very much so). Each participant evaluated eight different trajectories balanced using a Latin square design: four trajectories with gazing at the robot and four trajectories without gazing at the robot. 

Before the experiment, we collected data using the Spot robot's joystick controller to create the LfD module described in Section \ref{sec:lfd}, consisting of 18 expert demonstrations from 4 different people. 
We carefully collected data from experts with feedback from the target person, such as if the person felt safe or threatened according to the motion. 
The estimated reward function from the LfD is shown in Figure \ref{fig:irl}. The estimated reward increases when the robot gets closer to a target person, where the coordinate and the angle of the target person are fixed as $(0,0,0)$. On the learned reward function with gaze condition, the robot approaches from the front of the person, and the robot obtains a larger reward by heading to the target person at a decreasing speed. Under the no-gaze condition, the robot approaches from the back, and the reward function contains a higher value for heading diagonally to a target person. 


Figure \ref{fig:exp_app} shows the robotic motion of the proposed method and how the trajectories vary between different approaching methods. In Figure \ref{fig:exp_app}-(1), when the target person is facing the robot, the robot moves sideways in a short distance and goes straight to the person at a decreasing speed. When the target person is not looking at the robot, as shown in Figure \ref{fig:exp_app}-(2), it turns toward the direction in which the person is looking and then moves laterally to the target. Figure \ref{fig:exp_app}-(3) and \ref{fig:exp_app}-(4) illustrate the trajectories from different methods conditioned on the gaze parameter. The limitation of the knowledge distillation (KD) framework is that the robot does not slow down near the human. 
As the LfD-based reward function is trained from various complex expert demonstrations, it models diverse directions to approach motion, leading to frequent changes in heading directions. 
The proposed method tends to slowly approach from the forward direction with a slightly turning sideway on gaze condition. In no-gaze condition, the robot slowly approaches from sideways with lateral movement. 
We believe that the proposed hybrid learning-based approach has the benefit of smoothing the LfD-based trajectories, reducing the number of abrupt changes in the heading of the robot. 



%

We have conducted human-subject pilot studies using the RoSAS. To analyze our results, we conducted a two-way repeated-measures MANOVA on the RoSAS results obtained from the real-world approach experiment. As we have recruited limited subjects in this pilot, this study is not sufficiently powered to assess the effect;
the results should be interpreted with caution. We observe that the approaching motion significantly affects the participants on the \textit{warmth} dimension of the RoSAS ($F(7)=1.297$, $p=0.047$, $\eta^2 = 0.132$ with $\alpha = 0.05$). No other RoSAS dimension exhibits any significant results. Post hoc pairwise comparisons with Bonferroni correction were not able to detect any significant effects, likely due to a large number of contrasts (seven) as well as reduced statistical power. 

Table \ref{tab:rosas_1} reports the mean and standard deviation of each rating with different methods. The proposed method outperforms the other methods with a gap of $0.15$, $0.19$, and $0.27$, respectively, to each attribute. With curved motion and decrement in speed, the participants observed motion to be more competent and warm with less discomfort compared to straight and fast-approaching motion. Furthermore, we measure the difference between the robot-experienced group and the robot-inexperienced group in Table \ref{tab:rosas_2}. Exp denotes the robot-experienced group, while IExp. denotes the robot-inexperienced group.
The experienced group preferred the proposed method compared to the other method with a gap of $0.48$ while the in-experienced group preferred the baseline approach with a minor gap of $0.01$. Figure \ref{fig:rosas} reports the mean of each rating with different methods on gaze conditions. 
From the survey comments, two out of six people commented the lateral motion from the proposed method made it hard for the inexperienced group to predict the direction of the robot, while the experienced group observed that the motion was more dog-like. 
As such, we would like to carefully hypothesize that the expertise of the robot influences the preference of the motion of the robot when the person is not gazing at it. 

\subsection{Limitations}
Although the human-subject studies of the approaching motion did not appear to show significant effects, we would like to state that the user had a large variance in their perspectives. 
As the data of the learning from demonstration framework is collected from robot-experienced experts, the proposed method adapts the preferences of the experts more than the non-expert group. We believe that the proposed method can be further improved by considering the preference of the user. 
In addition, we have observed that the system fails on ambiguous commands or images during the search phase.

%
%
\section{Conclusion} 
In this paper, we have introduced \method, which tackles the problem of \textit{human search and approach} based on textual descriptions. We proposed the Human Search Socratic Model, which connects pre-trained models (e.g., large-language models and vision-language models) by sharing the language domain for solving the human search task. We evaluated the efficiency of the proposed search module in both real-world and simulation environments. The proposed method that its strength in performing a shorter path length with a small number of feedback from the user. Furthermore, we have presented a hybrid learning-based framework for generating the robotic motion to approach a person in a socially aware manner. We conducted pilot studies with human participants to analyze the properties of the proposed approach module to measure the user rate of three attributes. We observed that the proposed method participants outperformed the baseline approach method in all three attributes due to the slow and lateral movement of the proposed method. 
We believe that the proposed method can be further improved by considering the preferences of the user. 



\newpage
\bibliographystyle{unsrt}
\bibliography{references}

\begin{thebibliography}{10}

\bibitem{17_carpinella}
Colleen~M. Carpinella, Alisa~B. Wyman, Michael~A. Perez, and Steven~J.
  Stroessner.
\newblock The robotic social attributes scale (rosas): Development and
  validation.
\newblock In {\em Proc. of the 2017 12th ACM/IEEE International Conference on
  Human-Robot Interaction (HRI}, pages 254--262, 2017.

\bibitem{20_brown}
Tom Brown, Benjamin Mann, Nick Ryder, Melanie Subbiah, Jared~D Kaplan, Prafulla
  Dhariwal, Arvind Neelakantan, Pranav Shyam, Girish Sastry, Amanda Askell,
  et~al.
\newblock Language models are few-shot learners.
\newblock In {\em Proc. of the Advances in Neural Information Processing
  Systems (NeurIPS)}, volume~33, pages 1877--1901, 2020.

\bibitem{21_radford}
Alec Radford, Jong~Wook Kim, Chris Hallacy, Aditya Ramesh, Gabriel Goh,
  Sandhini Agarwal, Girish Sastry, Amanda Askell, Pamela Mishkin, Jack Clark,
  et~al.
\newblock Learning transferable visual models from natural language
  supervision.
\newblock In {\em Proc. of the International Conference on Machine Learning
  (ICML)}, pages 8748--8763. PMLR, 2021.

\bibitem{22_ahn}
Michael Ahn, Anthony Brohan, Noah Brown, Yevgen Chebotar, Omar Cortes, Byron
  David, Chelsea Finn, Keerthana Gopalakrishnan, Karol Hausman, Alex Herzog,
  et~al.
\newblock Do as i can, not as i say: Grounding language in robotic affordances.
\newblock {\em arXiv preprint arXiv:2204.01691}, 2022.

\bibitem{22_jang}
Eric Jang, Alex Irpan, Mohi Khansari, Daniel Kappler, Frederik Ebert, Corey
  Lynch, Sergey Levine, and Chelsea Finn.
\newblock Bc-z: Zero-shot task generalization with robotic imitation learning.
\newblock In {\em Proc. of the Conference on Robot Learning (CoRL)}, pages
  991--1002. PMLR, 2022.

\bibitem{20_zhu}
Fengda Zhu, Yi~Zhu, Xiaojun Chang, and Xiaodan Liang.
\newblock Vision-language navigation with self-supervised auxiliary reasoning
  tasks.
\newblock In {\em Proc. of the Conference on Computer Vision and Pattern
  Recognition (CVPR)}, pages 10012--10022, 2020.

\bibitem{22_shah}
Dhruv Shah, Blazej Osinski, Brian Ichter, and Sergey Levine.
\newblock Lm-nav: Robotic navigation with large pre-trained models of language,
  vision, and action.
\newblock In {\em Proc. of the Conference on Robot Learning (CoRL)}, 2022.

\bibitem{22_gadre}
Samir~Yitzhak Gadre, Mitchell Wortsman, Gabriel Ilharco, Ludwig Schmidt, and
  Shuran Song.
\newblock Clip on wheels: Zero-shot object navigation as object localization
  and exploration.
\newblock {\em arXiv preprint arXiv:2203.10421}, 2022.

\bibitem{22_khandelwal}
Apoorv Khandelwal, Luca Weihs, Roozbeh Mottaghi, and Aniruddha Kembhavi.
\newblock Simple but effective: Clip embeddings for embodied ai.
\newblock In {\em Proc. of the Conference on Computer Vision and Pattern
  Recognition (CVPR)}, pages 14829--14838, 2022.

\bibitem{22_majumdar}
Arjun Majumdar, Gunjan Aggarwal, Bhavika Devnani, Judy Hoffman, and Dhruv
  Batra.
\newblock Zson: Zero-shot object-goal navigation using multimodal goal
  embeddings.
\newblock {\em arXiv preprint arXiv:2206.12403}, 2022.

\bibitem{23_park}
Jeongeun Park, Taerim Yoon, Jejoon Hong, Youngjae Yu, Matthew Pan, and Sungjoon
  Choi.
\newblock Zero-shot active visual search (zavis): Intelligent object search for
  robotic assistants.
\newblock In {\em Proc. of the IEEE International Conference on Robotics and
  Automation (ICRA)}. IEEE, 2023.

\bibitem{22_zeng}
Andy Zeng, Adrian Wong, Stefan Welker, Krzysztof Choromanski, Federico Tombari,
  Aveek Purohit, Michael Ryoo, Vikas Sindhwani, Johnny Lee, Vincent Vanhoucke,
  et~al.
\newblock Socratic models: Composing zero-shot multimodal reasoning with
  language.
\newblock In {\em Proc. of the International Conference on Learning
  Representations (ICLR)}, 2022.

\bibitem{22_huang}
Wenlong Huang, Fei Xia, Ted Xiao, Harris Chan, Jacky Liang, Pete Florence, Andy
  Zeng, Jonathan Tompson, Igor Mordatch, Yevgen Chebotar, et~al.
\newblock Inner monologue: Embodied reasoning through planning with language
  models.
\newblock In {\em Proc. of the Conference on Robot Learning (CoRL)}, 2022.

\bibitem{16_selvaraju}
Ramprasaath~R Selvaraju, Michael Cogswell, Abhishek Das, Ramakrishna Vedantam,
  Devi Parikh, and Dhruv Batra.
\newblock Grad-cam: Visual explanations from deep networks via gradient-based
  localization.
\newblock In {\em Proc. of the IEEE international conference on computer vision
  (ICCV)}, pages 618--626, 2017.

\bibitem{98_yamauchi}
Brian Yamauchi.
\newblock Frontier-based exploration using multiple robots.
\newblock In {\em Proc. of the International conference on Autonomous agents},
  pages 47--53, 1998.

\bibitem{17_ahn}
Hyemin Ahn, Yoonseon Oh, Sungjoon Choi, Claire~J Tomlin, and Songhwai Oh.
\newblock Online learning to approach a person with no regret.
\newblock {\em IEEE Robotics and Automation Letters}, 3(1):52--59, 2017.

\bibitem{21_konar}
Abhisek Konar, Bobak~H Baghi, and Gregory Dudek.
\newblock Learning goal conditioned socially compliant navigation from
  demonstration using risk-based features.
\newblock {\em IEEE Robotics and Automation Letters}, 6(2):651--658, 2021.

\bibitem{20_chen}
Yuying Chen, Congcong Liu, Bertram~E Shi, and Ming Liu.
\newblock Robot navigation in crowds by graph convolutional networks with
  attention learned from human gaze.
\newblock {\em IEEE Robotics and Automation Letters}, 5(2):2754--2761, 2020.

\bibitem{22_tiong}
Anthony Meng~Huat Tiong, Junnan Li, Boyang Li, Silvio Savarese, and Steven~CH
  Hoi.
\newblock Plug-and-play vqa: Zero-shot vqa by conjoining large pretrained
  models with zero training.
\newblock In {\em Proc. of the Empirical Methods in Natural Language Processing
  (EMNLP)}, 2022.

\bibitem{13_mikolov}
Tomas Mikolov, Kai Chen, Greg Corrado, and Jeffrey Dean.
\newblock Efficient estimation of word representations in vector space.
\newblock In {\em Proc. of the International Conference on Learning
  Representations (ICLR)}, 2013.

\bibitem{2020_jocher}
Glenn Jocher.
\newblock {ultralytics/yolov5: v3.1 - Bug Fixes and Performance Improvements}.
\newblock \url{https://github.com/ultralytics/yolov5}, October 2020.

\bibitem{14_lin}
Tsung-Yi Lin, Michael Maire, Serge Belongie, James Hays, Pietro Perona, Deva
  Ramanan, Piotr Doll{\'a}r, and C~Lawrence Zitnick.
\newblock Microsoft coco: Common objects in context.
\newblock In {\em Proc. of the European conference on computer vision (ECCV)},
  pages 740--755. Springer, 2014.

\bibitem{19_lugaresi}
Camillo Lugaresi, Jiuqiang Tang, Hadon Nash, Chris McClanahan, Esha Uboweja,
  Michael Hays, Fan Zhang, Chuo-Ling Chang, Ming~Guang Yong, Juhyun Lee, et~al.
\newblock Mediapipe: A framework for building perception pipelines.
\newblock {\em arXiv preprint arXiv:1906.08172}, 2019.

\bibitem{22_li}
Junnan Li, Dongxu Li, Caiming Xiong, and Steven Hoi.
\newblock Blip: Bootstrapping language-image pre-training for unified
  vision-language understanding and generation.
\newblock In {\em Proc. of the International Conference on Machine Learning
  (ICML))}, 2022.

\bibitem{16_choi}
Sungjoon Choi, Kyungjae Lee, Andy Park, and Songhwai Oh.
\newblock Density matching reward learning.
\newblock {\em arXiv preprint arXiv:1608.03694}, 2016.

\bibitem{janson2015fast}
Lucas Janson, Edward Schmerling, Ashley Clark, and Marco Pavone.
\newblock Fast marching tree: A fast marching sampling-based method for optimal
  motion planning in many dimensions.
\newblock {\em The International journal of robotics research}, 34(7):883--921,
  2015.

\bibitem{sucan2012ompl}
Ioan~A. {\c{S}}ucan, Mark Moll, and Lydia~E. Kavraki.
\newblock The {O}pen {M}otion {P}lanning {L}ibrary.
\newblock {\em {IEEE} Robotics \& Automation Magazine}, 19(4):72--82, December
  2012.
\newblock \url{https://ompl.kavrakilab.org}.

\bibitem{19_zhang}
Ying Zhang, Guohui Tian, Jiaxing Lu, Mengyang Zhang, and Senyan Zhang.
\newblock Efficient dynamic object search in home environment by mobile robot:
  A priori knowledge-based approach.
\newblock {\em IEEE Transactions on Vehicular Technology}, 68(10):9466--9477,
  2019.

\bibitem{21_oakd}
OpenCV~Luxonis (n.d.).
\newblock Opencv ai kit, July 2021.

\bibitem{18_anderson}
Peter Anderson, Angel Chang, Devendra~Singh Chaplot, Alexey Dosovitskiy,
  Saurabh Gupta, Vladlen Koltun, Jana Kosecka, Jitendra Malik, Roozbeh
  Mottaghi, Manolis Savva, et~al.
\newblock On evaluation of embodied navigation agents.
\newblock {\em arXiv preprint arXiv:1807.06757}, 2018.

\bibitem{09_kollar}
Thomas Kollar and Nicholas Roy.
\newblock Utilizing object-object and object-scene context when planning to
  find things.
\newblock In {\em 2009 IEEE International Conference on Robotics and
  Automation}, pages 2168--2173, 2009.

\bibitem{22_minderer}
Matthias Minderer, Alexey Gritsenko, Austin Stone, Maxim Neumann, Dirk
  Weissenborn, Alexey Dosovitskiy, Aravindh Mahendran, Anurag Arnab, Mostafa
  Dehghani, Zhuoran Shen, et~al.
\newblock Simple open-vocabulary object detection with vision transformers.
\newblock In {\em Proc. of the European Conference on Computer Vision (ECCV)},
  2022.

\bibitem{Zeevi_97}
Assaf~J Zeevi and Ronny Meir.
\newblock Density estimation through convex combinations of densities:
  approximation and estimation bounds.
\newblock {\em Neural Networks}, 10(1):99--109, 1997.

\bibitem{ho2016generative}
Jonathan Ho and Stefano Ermon.
\newblock Generative adversarial imitation learning.
\newblock {\em Advances in neural information processing systems}, 29, 2016.

\bibitem{gilks2005markov}
Walter~R Gilks, Sylvia Richardson, and David Spiegelhalter.
\newblock {\em Markov chain Monte Carlo in practice}.
\newblock CRC press, 1995.

\bibitem{Pollard_00}
David Pollard.
\newblock Asymptopia.
\newblock {\em Manuscript, Yale University, Dept. of Statist., New Haven,
  Connecticut}, 2000.

\bibitem{Alfeld_04}
P~Alfeld.
\newblock Understanding mathematics. utah: Departemen of mathematics, 2014.

\end{thebibliography}

\onecolumn
\section*{Appendix}

\subsection{Failure Cases}
\begin{figure}[!h]
    \centering
    \includegraphics[width=0.8\textwidth]{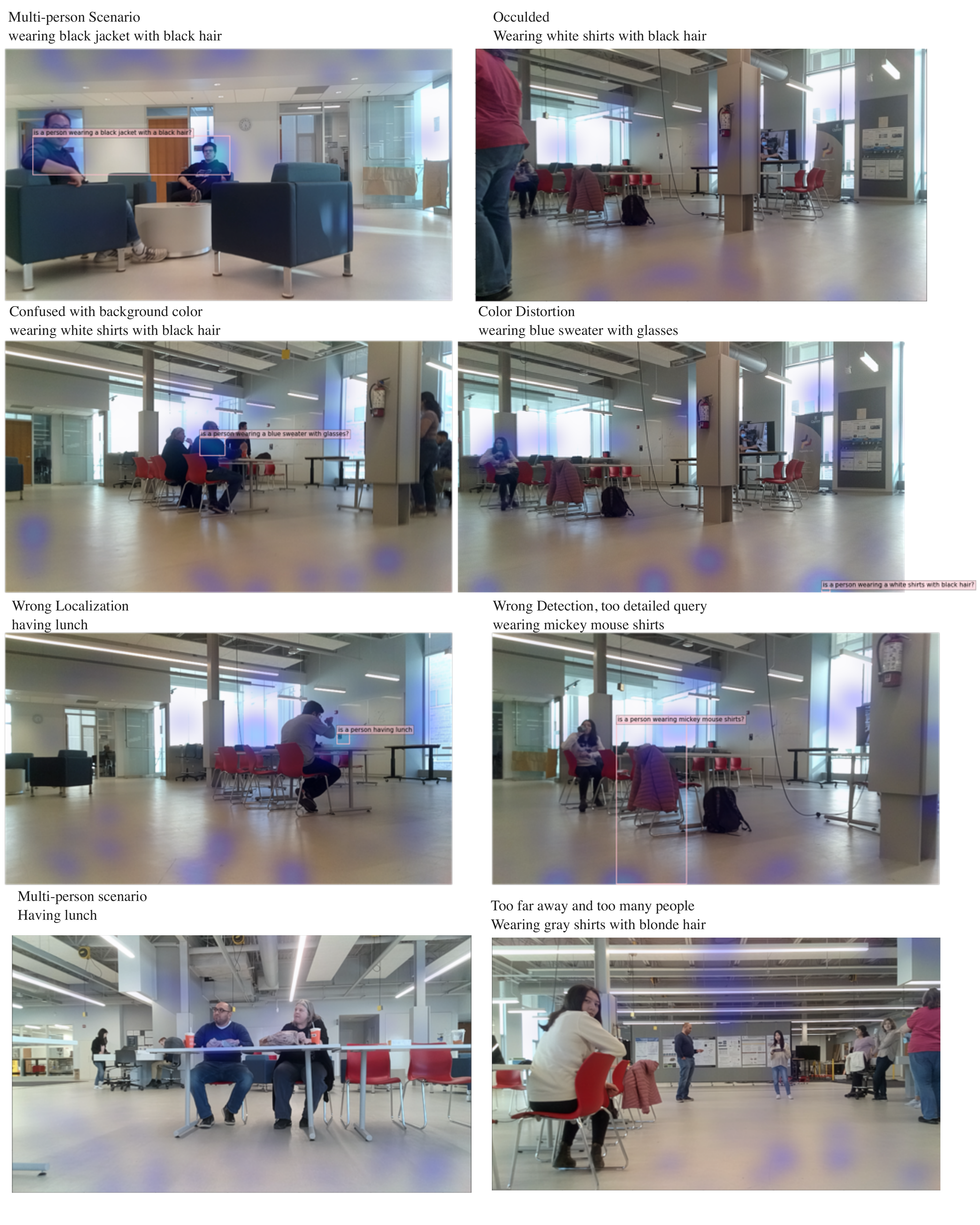}
    \caption{Failure Cases on Vision}
    \label{fig:det_fail}
\end{figure}

We observe that the system may fail when presented with ambiguous or noisy inputs. In the vision system, localizing a person using grad-CAM may fail in the presence of multiple people in one scene that corresponds to the prompt. Additionally, the system may be susceptible to failures in certain lighting conditions or when the person's clothing color is similar to the background color. Moreover, because the environment is dynamic, occlusion by other people in the scene can also lead to search failure. The failure cases are illustrated in Figure \ref{fig:det_fail}. 

In addition, ambiguous location clues can also cause system failures, for example, the term "teacher" may refer to either a student teaching assistant or a professor. Furthermore, queries that are out-of-distribution, such as "faculty members" who usually stay at a place not annotated, tend to result in incorrect predictions. The search prior on the ambiguous commands are illustrated in Table \ref{tab:amb_search}

\begin{table}[!h]
    \centering
    \begin{tabular}{|m{4cm}|m{1cm}|m{1cm}|m{1cm}|m{1cm}|m{1cm}|}
        \hline
        Location Clue & (1) & (2) & (3) & (4) & (5) \\
        \hline
        teacher & 	0.694 &	0.931	& 0.1	& 0.986 &	0.93 \\ 
        \hline
teaching assistant& 	0.95	& 0.84 & 	0.3& 	1.0& 	1.0 \\ 
\hline
faculty member & 	0.55	& 0.9& 	0	& 0.99	& 0.95 \\ 
\hline
faculty member and having lunch	& 0.4	& 0.95	& 0	& 0.93 &0.80 \\
\hline
student who is walking near chairs &  0.95 & 0.52 & 0.10 & 0.94 & 0.60 \\
\hline
cleaning staff & 0.58 & 0.55 & 0.15 & 0.93 & 0.84 \\
\hline
intern & 0.95 & 0.73 & 0.15 & 1.0 & 1.0 \\
\hline
undergraduate intern & 0.9 & 0.58 & 0.1 & 0.98 & 0.93 \\
\hline
    \end{tabular}
    \caption{Search prior with ambiguous commands. (1) Desk for students (2) Office for professors (3) Sofa (4) Open space desk (5) Workspace with robots}
    \label{tab:amb_search}
\end{table}

\subsection{Various Prompts}
\begin{figure}[!h]
    \centering
    \includegraphics[width=0.8\textwidth]{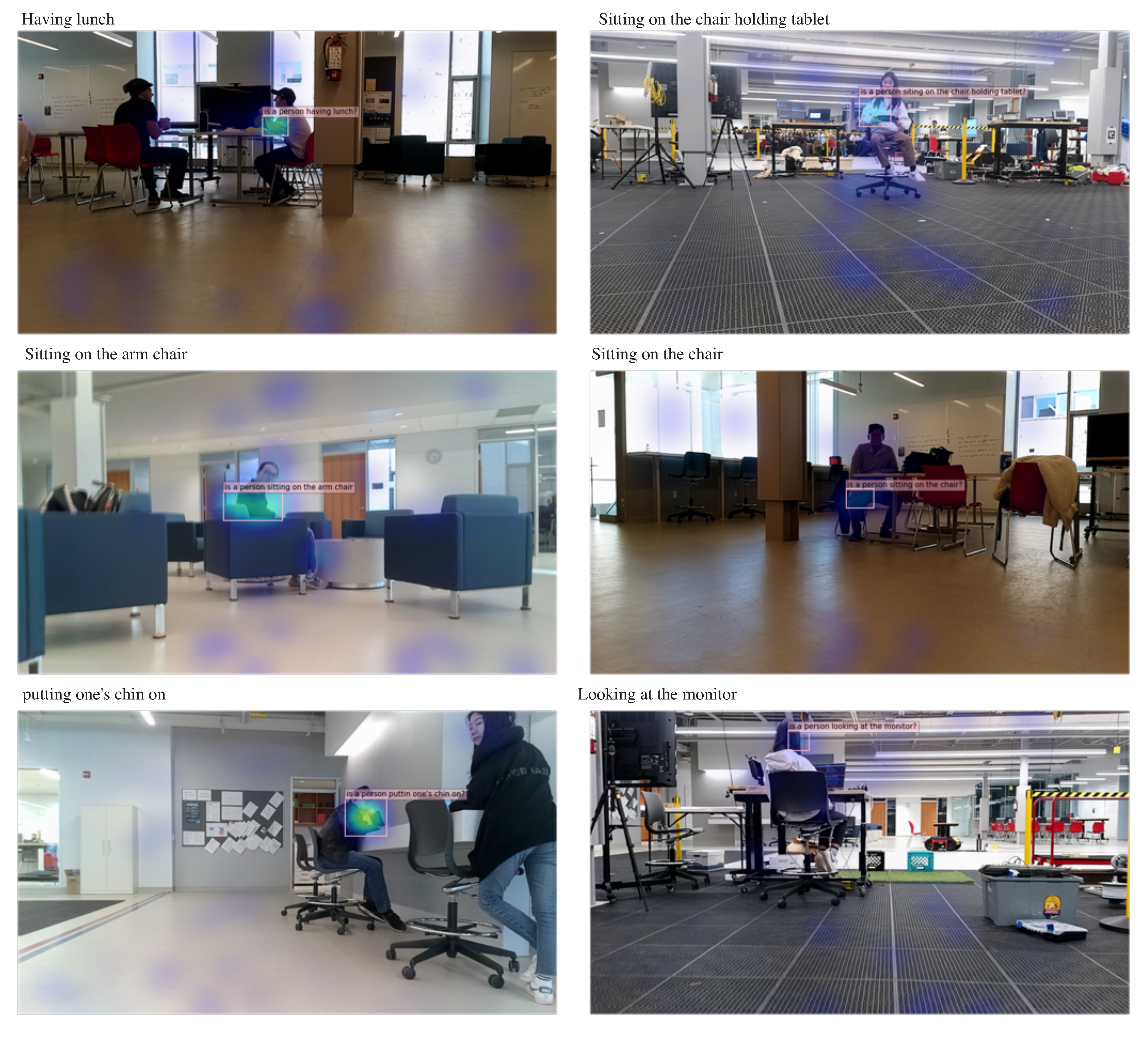}
    \caption{Various Prompts on Vision}
    \label{fig:det_var}
\end{figure}

We have conducted additional experiments on Language clarifying the generalizability of our system to more complex queries. The prompt used in LLM is simply a set of the categorical names, e.g., desk, sofa, office, workspace, etc, of the annotated map, which does not explicitly include information about which user belongs to that area. However, the location clue from the input sentence, e.g., a student who researches AI, does not explicitly express the location of the target person. The possible location of the target person cannot be easily determined by simple heuristics, and we leveraged the commonsense knowledge of the large language model to address this challenge. 
We tested both the vision-language model and the large-language model on the complex queries. The illustrations of the results are shown in Figure \ref{fig:det_var} and Table \ref{tab:search_complex} respectively.

\begin{table}[!h]
    \centering
    \begin{tabular}{| m{11em} | m{0.7cm}| m{0.7cm} |m{0.7cm}|m{0.7cm}|m{0.7cm}|m{0.7cm}|}
        \hline
        Location Clue & (1) & (2) & (3) & (4) & (5)& (6) \\
        \hline
        student from mechanical engineering department & 1.0  &  0.35 & 0.15 & 0.25 & 0.7 & 0.5 \\
        \hline
        student from the computer science department & 1.0  & 0.0  &  0.0  & 0.0 & 0.6 & 0.35  \\
        \hline
        assistant professor & 0.6  & 0.05 & 0.05 & 0.97& 0.6 & 0.15    \\
        \hline
        professor & 0.14 & 0.05 & 0.05 & 0.98 & 0.1 & 0.05\\
        \hline
        student working on the paper & 1.0 & 0.25 & 0.14 & 0.15 & 0.75 & 0.45  \\
        \hline
        visiting research students to do experiments & 1.0 & 0.15 & 0.50 & 0.31 & 0.9 & 0.65 \\
        \hline
    \end{tabular}
    \caption{Search prior with complex commands. (1) Desk for students (2)Machine shop (3) Sofa (4) Professor Office (5) Open space desk (6) Workspace with robots}
    \label{tab:search_complex}
\end{table}
 
\subsection{Word Based Path Generation Function}
The algorithm for generating the reward function based on knowledge distillation is mentioned in Section \ref{sec:kd}. To generate a trajectory based on a set of keywords $\{w^l_i\}_{l=1}^L=f_k(s^t_i)$, we set a word-to-trajectory function $f_d(w)$ for generating a partial path based on the word. The word-to-trajectory function function $f_d(w)$ is as follows:
\begin{equation}
	f_d(w) = 
	\begin{cases}
	\{(0,\ell \cos (\theta_r^h)), (0,\ell \sin (\theta_r^h))\}, & \text{if }  w = \text{straight}\\
    \{ (0,\ell \cos (\pi/4) \text{sgn} (x_r^h)), \\
            (0,\ell \sin (\pi/4) \text{sgn} (x_r^h))\}, \\
    \{ (0,\ell \cos (\pi/4) \text{sgn} (x_r^h)), \\
            (0,\ell \sin (-\pi/4) \text{sgn} (x_r^h))\}
            & \text{if }  w = \text{45}\\
    \{ (0,\ell \cos (\pi/2), (0,\ell \sin (\pi/2))\}, \\
	\{ (0,\ell \cos (\pi/2), (0,\ell \sin (-\pi/2))\}
            & \text{if }  w = \text{side}\\
        \{ \{\ell/2 + \ell \cos(\theta)\}_{\theta = 0}^{\pi}\text{sgn} (x_r^h), \\
            \{\ell/2 + \ell \sin(\theta)\}_{\theta = 0}^{\pi}\text{sgn} (x_r^h)\}, \\
	\{ \{\ell/2 + \ell \cos(\theta)\}_{\theta = -\pi}^{0}\text{sgn} (x_r^h), \\
            \{\ell/2 + \ell \sin(\theta)\}_{\theta = -\pi}^{0}\text{sgn} (x_r^h)\},)\}
            & \text{if }  w = \text{curved or curve} \\
        \{(0,\ell)), (0)\}, & \text{if }  w = \text{front}\\
        \{(0,-\ell)), (0)\}, & \text{if }  w = \text{behind}\\
	\end{cases}
 \label{eq:kd_dir}
\end{equation}
where $\ell$ is a step variable. We set the step variable on the experiments as $\ell=0.2$.

\subsection{Function for Determining the Visit of Waypoint}
On the local search phase of the waypoint generator on Section \ref{sec:search_wg}, we utilize the history of waypoints to guide the robot to visit waypoints that have not been previously visited. The equation to determine the robot to visit the waypoint is as follows:
\begin{equation}
	go(\mathbf{p}_r) = 
	\begin{cases}
	1, & \text{if }  \min_i ||\mathbf{p}_r - \mathbf{p}_i||_2 > t_g \\
	0, & \text{otherwise}
	\end{cases}
\end{equation}
where $t_g$ is a distance threshold to determine whether the robot has visited the waypoint or not. 

\subsection{Experimental Setup}
 In the search phase, we experimented on a 2-floor lab environment with 40 experiments in total, and the details are shown in URL section 4. In addition, in the approach phase, we had 16 participants with each evaluating 8 trajectories, which leads to 128 experiments in total.
\begin{table}[!h]
    \centering
    \begin{tabular}{|c|c|c|c|c|c|}
        \hline
         Environment & \# Q & Trials & \# A & Methods & Total \\
         \hline
         Real-world Lab (2-floors)& 5 & 2 & 5 & 4 & 40 \\
         \hline
         Simulation Household & 4 & 2 & 3 & 4 & 32 \\
         \hline
         Simulation Office & 8 & 2 & 3 & 4 & 64\\
         \hline
         Total & 17 & & & & 168 \\ 
         \hline
    \end{tabular}
    \caption{Search Experiment Setup. \# Q denotes the number of input queries. Trials denote different initial positions, \# A denotes the number of categories in annotated map. Methods denote the number of compared methods. }
    \label{tab:setup}
\end{table}

\subsection{Hyperparameters} \label{sec:appendix_hyp}
We set the number of generation $M$ for prior knowledge on equation \ref{eq:prior} as $20$, and the weight of prior knowledge $w_e$ on equation \ref{eq:cost_search} as $30$. In addition, we set the threshold for the detection $t_c$ as $0.5$. Finally, we choose the threshold of the visitence of the waypoint $t_g$ as $2$.

For the KDMRL framework on Section \ref{sec:lfd}, we set hyperparamters as follows: $\beta = 0.2$, $Z=200$, $\delta = 0.8$, $\lambda=0.01$. We set the weight for the combined reward function as $w_r=0.2$. We leveraged the radial basis kernel function with $\sigma$ as $1.0, 0.5, 1.0$ for $k(\cdot, \cdot), k_{\mu}(\cdot, \cdot), k_r(\cdot, \cdot)$ respectively. On the path planning module, we set $w_p = 1, w_o = 0.5, \zeta = 1.5$ on equations \ref{eq:approach_cost1} and \ref{eq:approach_cost2}. 

\subsection{Prompts of the Search Experiments}\label{sec:appendix_prompts}
Table \ref{tab:env} shows the prompts used in search experiments.
\begin{table}[h]
    \centering
    \begin{tabular}{c|c}
         Appearance ($t_1$) & Location Clue ($t_2$)\\
         \hline \hline
         \multicolumn{2}{c}{Real-World} \\ 
         \hline
         wearing a white shirt with black hair & student that works with robots\\
         wearing a black shirt with black hair & student that works with robots \\
         wearing a black shirt with a hat & professor \\
         wearing a red knit with black hair & student that works on VR \\
         wearing a white shirt with black hair & student that works on AI \\
         \hline \hline
         \multicolumn{2}{c}{Simulation} \\
         \hline
         wearing a white shirt with black hair & dad and cooking for dinner \\
         wearing a blue shirt with white hair & mom and cooking for dinner \\
         wearing a striped shirt with black hair & kids and preparing for bed \\ 
         wearing a striped shirt with brown hair & grandma and watching TV \\ 
        wearing a yellow shirt with brown hair & employee and doing office work \\
        wearing a green shirt with black hair & employee and taking a rest \\
        wearing a purple shirt with black hair & employer and doing a meeting \\ 
        wearing a yellow shirt with blonde hair & employee and doing office work \\
        wearing a red shirt with brown hair & employee and doing office work \\
        wearing a purple shirt with brown hair & employee and doing a meeting \\
        wearing a orange shirt with blonde hair & employee and doing office work \\
        wearing a blue shirt with gray hair & staff and taking a rest \\ 
        \hline
    \end{tabular}
    \caption{Prompts of the experimental environment}
    \label{tab:env}
\end{table}

\subsection{Vision-Language Detection}
Following the text-based localization of the image in Section \ref{sec:vlm}, we localize the target person based on un-normalized Grad-CAM on the text and image similarity. We obtain the bounding box $\mathbf{b} = \{x_1,y_1,x_2,y_2\}$ of the image with threshold $t_c$ as follows: 
\begin{equation}
	x_1, y_1 = \arg \min_{x,y} L^{\text{text}}_{\text{Grad-CAM}}(x,y) > t_c
\end{equation}

\begin{equation}
	x_2, y_2 = \arg \max_{x,y} L^{\text{text}}_{\text{Grad-CAM}}(x,y) > t_c 
\end{equation}


\subsection{Other Vision Model (OWL-ViT)}
\begin{figure}[!h]
    \centering
    \includegraphics[width=0.8\textwidth]{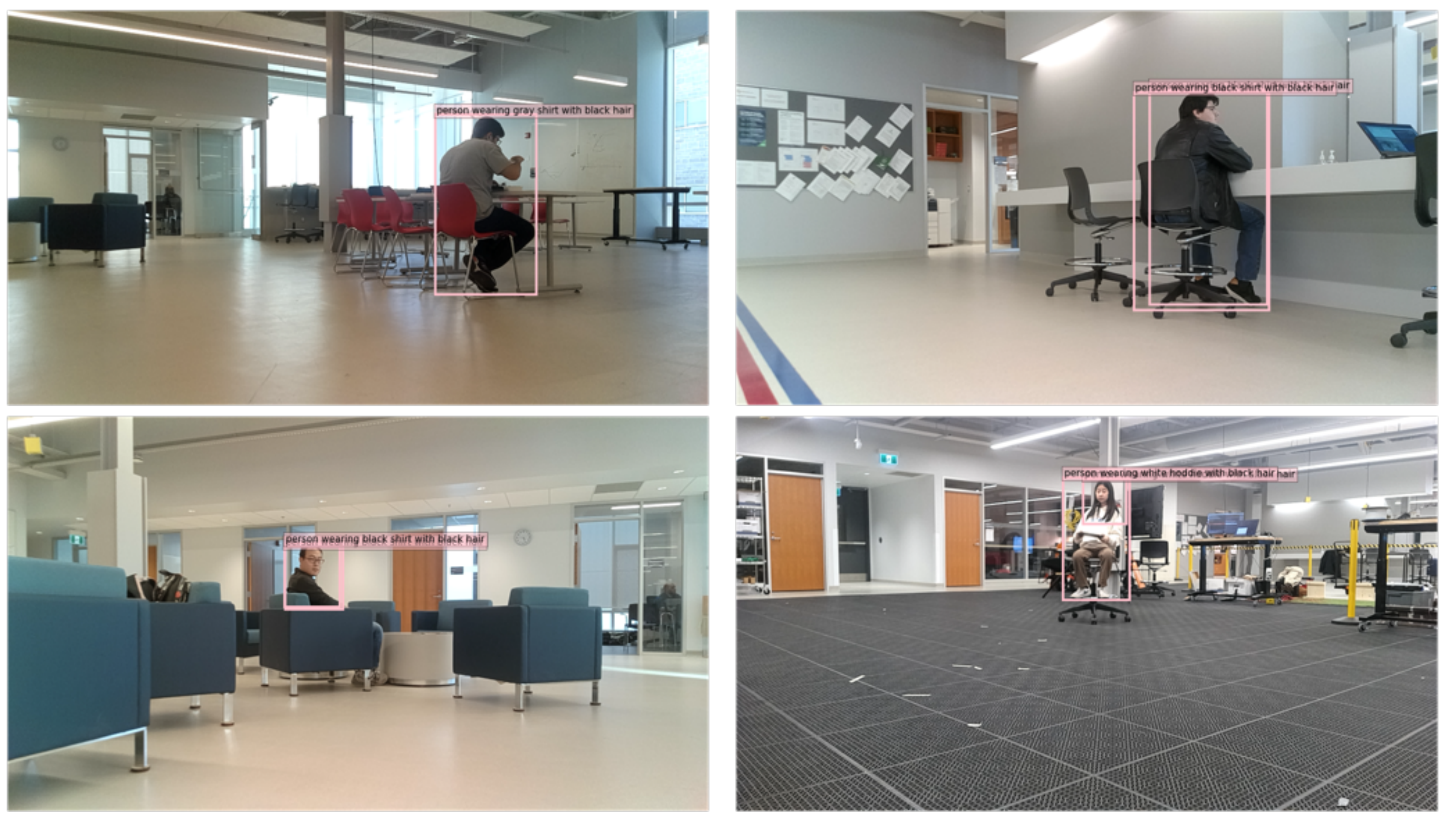}
    \caption{Results with Owl-Vit}
    \label{fig:det_owl}
\end{figure}

For the vision-language model, we had two objectives; localizing the target person and obtaining a textual description of the scene. As we used a laptop with limited computational power to control the Spot robot, we could not use both the detection model and captioning model on one GPU. We have also tested the proposed method on the Open-Vocabulary detection model, in particular, owl-vit \cite{22_minderer}. The results are shown in Figure \ref{fig:det_owl}.

\subsection{Theoretical Analysis of KDMRL}
In this section, we present a bound on the sample complexity of the DMRL mentioned in Section \ref{sec:lfd}. Following the objective of the inverse reinforcement learning formulation, the objective of density matching reward learning (DMRL) is as follows:
\begin{equation}
\begin{split}
	\underset{R}{\text{maximize}} \  & V(R) = <\hat{\mu}, R> \\
	\text{subject to}\  & ||R||_2 \leq 1
\end{split}
\label{eq:kdmrl_obj}
\end{equation}
where $\hat{\mu}$ is an estimated density of the state space. 

Suppose that we have access to the true stationary state-action density
$\bar{\mu}(s, a)$ and estimate the reward $\bar{R}(s, a)$ from (\ref{eq:kdmrl_obj}),
i.e., $\bar{R} = \arg \max_{ ||R||_2 \le 1} \langle \bar{\mu}, R \rangle $.
We define the value of $\bar{R}$ with respect to the true state-action density
$\bar{\mu}(s, a)$ as an optimal value $\bar{V} = \langle \bar{\mu}, \bar{R} \rangle$.
In other words, the optimal value $\bar{V}$ is an expectation of reward $\bar{R}$
with respect to the true state-action density $\bar{\mu}$, i.e.,
\begin{equation} \label{eq:optv}
	\bar{V} = \langle \bar{\mu}, \bar{R} \rangle.
\end{equation}

Similarly, we define an estimated value $\hat{V}$ as an expectation of an estimated reward $\hat{R}$ computed from solving  (\ref{eq:kdmrl_obj}) with estimated density $\hat{\mu}$ with respect to the true state action density
$\bar{\mu}(s, a)$, i.e.,
\begin{equation} \label{eq:estv}
	\hat{V} = \langle \bar{\mu}, \hat{R} \rangle.
\end{equation}

As estimating the true density $\bar{\mu}(s, a)$ is practically infeasible, as such it requires an infinite number of samples, we give a probabilistic bound on the absolute difference between $\bar{V}$ and $\hat{V}$, i.e., $| \bar{V} - \hat{V} |$ as a function of the number of trajectories (sample complexity). We show the sample complexity on the absolute difference between the best value $\bar{V}$  and the optimized value, $\hat{V}$, with $n$ samples. As $| \bar{V} - \hat{V} | = | \langle \bar{\mu}, \, \bar{R}-\hat{R} \rangle |$,  minimizing this quantity can be interpreted as minimizing the difference between the estimated reward $\hat{R}$ and the true reward $\bar{R}$ when projected by the true state-action density $\bar{\mu}$. 

We first introduce assumptions for the remaining theorem. Note that assumption \ref{ass:1} to \ref{ass:7} are illustrated in \cite{Zeevi_97}, but we depict the assumptions here for completeness of the theorem. Assumption \ref{ass:8} is newly added. 

\begin{assumption}{}{}\label{ass:1}
	The random variables $\{ x_i\}_i^N$ are sampled 
	independent and identically distributed
	according to $f(x)$. 
\end{assumption}
\begin{assumption}{}{}\label{ass:2}
	The density estimator function $f_n^{\theta}$ is piecewise continuous 
	for each parameter $\theta$.
\end{assumption}
\begin{assumption}{}{}\label{ass:3}
	(a) $E[\log f(x)]$ exists and $|\log f_n^{\theta}(x)| \le m(x) ~ \forall \theta \in \Theta$
	for some $m(x)$
	where $m(x)$ is an integrable function with respect to $f$.
	(b) $E[\log(f/f_n^{\theta})]$ has a unique minimum. 
\end{assumption}
\begin{assumption}{}{}\label{ass:4}
	$\frac{\partial \log f_n^{\theta}(x)}{\partial \theta}$ 
	is integrable with respect to $x$
	and continuously differentiable. 
\end{assumption}
\begin{assumption}{}{}\label{ass:5}
	$|\frac{\partial^2 \log f_n^{\theta}(x)}{\partial \theta_i \partial \theta_j}|$
	and 
	$|\frac{\partial f_n^{\theta}(x)}{\partial \theta_i }\frac{\partial f_n^{\theta}(x)}{\partial \theta_j }|$
	is dominated by functions integrable with respect to $f$ $\forall x \in \mathcal{X}, ~ \theta \in \Theta$.
\end{assumption}
\begin{assumption}{}{}\label{ass:6}
	(a) $\theta^*$ is interior to $\Theta$. (b) $B(\theta^*)$ is nonsingular. 
	(c) $\theta^*$ is a regular point of $A(\theta)$. 
\end{assumption}
\begin{assumption}{}{}\label{ass:7}
	The convex model $f_n^{\theta} = \mathcal{G}_n$ obeys $\eta$ positivity requirement. 
\end{assumption}
\begin{assumption}{}{}\label{ass:8}
	The target density function $f \in \{ \sum_{i=1}^{\infty} \alpha_i \phi_{\rho}(\cdot ; \theta_i)  \}$
	where $\phi_{\rho}(\cdot ; \theta_i)$ is a basis density function.
\end{assumption}

The followings are the main results of this analysis. 
\begin{theorem} \label{thm:main1}
	Let $\hat{R}$ be computed from 
	$\hat{R} = \underset{R}{\mathrm{arg\max} \langle \hat{\mu}, R \rangle} \ \text{s.t.} \ \| R \|_2 \le 1$
	and $\bar{R}$ be computed from 
	$\bar{R} = \underset{R}{\mathrm{arg\max} \langle \bar{\mu}, R \rangle} \ \text{s.t.} \ \| R \|_2 \le 1$,
	where $\hat{\mu}$ and $\bar{\mu}$ are estimated and true
	densities, respectively. 
	Then
	\begin{equation*}
		| \langle \bar{\mu}, \hat{R} \rangle - \langle \bar{\mu}, \bar{R} \rangle | 
		\le
		3(R_{max} - R_{min})d_{var}(\bar{\mu}, \hat{\mu})
	\end{equation*}
	where $d_{var}(\cdot, \cdot)$ is the variational distance\footnote{
		The variational distance between two probability
		distributions, $P$ and $Q$, is defined as
		$d_{var}(P, Q) = \frac{1}{2}\sum_{x \in \mathcal{X}}|P(x)-Q(x)|$.
		It is also known that 
		$d_{var}(P, Q) = \sum_{x \in \mathcal{X}:P(x)>Q(x)}|P(x)-Q(x)|$.
		}
		between two probability distributions. 
\end{theorem}

\begin {theorem} \label{thm:main2}
	Suppose Assumption \ref{ass:1}--\ref{ass:8} hold.
	Let $n$ and $N$ be the number of samples and the number of	basis functions for kernel density estimation, respectively. Then, for any $\delta \ge 0$ and $n$ and $N$ be sufficiently large, we have
	\begin{equation}
		|\langle \bar{\mu}, \hat{R} \rangle - \langle \bar{\mu}, \bar{R} \rangle| 
		\le 3\sqrt{2} (R_{max}-R_{min}) \sqrt{ O\left( \frac{1}{n} \right) 
			+ O\left( \frac{nd}{N}(1+\frac{1}{\sqrt{\delta}}) \right) }
	\end{equation}
	with probability at least $(1-\delta)$.
\end {theorem}

The main intuition behind the DMRL is that IRL can be viewed as a dual problem of finding a reward function that matches  the state-action distribution \cite{ho2016generative}. The domain of our interest is the product space of 
states and actions, $\mathcal{S} \times \mathcal{A}$, and for the notational simplicity, we will denote this space as  $\mathcal{X} = \mathcal{S} \times \mathcal{A}$. Before proving Theorem \ref{thm:main1} and \ref{thm:main2},
we first introduce useful lemmas. 

\begin{lemma} {\cite{gilks2005markov}}
	\label{lem:1}
	Let $P$ and $Q$ be probability distributions over $\mathcal{X}$
	and $f$ be a bounded function on $\mathcal{X}$.
	Then,
	\begin{equation*}
		|\langle P, f \rangle - \langle Q, f \rangle| \le (\sup f - \inf f)d_{var}(P, Q).
	\end{equation*}
\end{lemma}

\begin{lemma} {\cite{Pollard_00}}
	\label{lem:2}
	Suppose $P$ and $Q$ are probability distributions. 
	Then, 
	\begin{equation*}
		d_{var}(P, \, Q) \le \sqrt{2} d_{\mathcal{H}}(P, \, Q),
	\end{equation*}
	where $d_{\mathcal{H}}(\cdot)$ is the Hellinger distance\footnote{
		The Hellinger distance between two probability distributions $P$ and $Q$
		is defined as $d^2_{\mathcal{H}}(P, Q) = \frac{1}{2}
			\sum_{x \in \mathcal{X}} (\sqrt{P(x)} - \sqrt{Q(x)})^2
		$
	}.
\end{lemma}

\begin{lemma}{\cite{Alfeld_04}}\label{lem:3} 
	Suppose $\| \cdot \|$ is a proper norm defined on index set $\mathcal{X}$.
	Then, for any $a, b \in \mathcal{X} $,
	\begin{equation*}
		|\| a \| - \| b \|| \le \| a-b \|.
	\end{equation*}
\end{lemma}

We also introduce a theorem from \cite{Zeevi_97}, which is used to prove Theorem \ref{thm:main2}. 
\begin{theorem} \label{thm:Zeevi_97}
	Suppose Assumption \ref{ass:1}--\ref{ass:8} hold.
	Let $n$ and $N$ be the number of samples and basis functions for
	kernel density estimation, respectively, and $\hat{f}_{n, N}$
	be the estimated density function from kernel density estimation 
	with $n$ samples and $N$ basis functions. 	
	Then, for any $\delta \ge 0$ and $n$ and $N$ sufficiently large,
	we have
	\begin{equation*}
		d^2_{\mathcal{H}}(f, \hat{f}_{n,N}) 
			\le O\left( \frac{1}{n} \right) 
			+ O\left( \frac{nd}{N}(1+\frac{1}{\sqrt{\delta}}) \right),
	\end{equation*}
	where $d^2_{\mathcal{H}}(\cdot)$ is a squared Hellinger distance between
	two probability distributions. 
\end{theorem}

With that Lemma \ref {lem:1}, \ref{lem:2}, \ref{lem:3} and Theorem \ref{thm:Zeevi_97}, we can prove Theorem \ref{thm:main1} and Theorem \ref{thm:main2}.

\begin{proof}(Theorem \ref{thm:main1})
	Let $\bar{\mu}$ and $\hat{\mu}$ be the true and estimated density functions, 
	respectively, and $\bar{R}$ and $\hat{R}$ be the reward functions
	estimated by DMRL with $\bar{\mu}$ and $\hat{\mu}$, respectively. 
	The absolute difference between the optimal value $\bar{V}$ and 
	the estimated value $\hat{V}$ can be expressed as:
	\begin{equation*}
		\begin{aligned} 
		&& |\langle \bar{\mu}, \hat{R} \rangle - \langle \bar{\mu}, \bar{R} \rangle| 
			& = |\langle \bar{\mu}, \hat{R} \rangle - \langle \bar{\mu}, \bar{R} \rangle
				+ \langle \hat{\mu}, \hat{R} \rangle - \langle \hat{\mu}, \hat{R} \rangle| \\
		&&  & \le |\langle \bar{\mu}, \hat{R} \rangle - \langle \hat{\mu}, \hat{R} \rangle |
				+ |\langle \bar{\mu}, \bar{R} \rangle - \langle \hat{\mu}, \hat{R} \rangle| \\
		&&  & \le (R_{max}-R_{min}) d_{var}(\bar{\mu}, \, \hat{\mu}) 
			+ | \langle \bar{\mu}, \, \bar{R} \rangle - \langle \hat{\mu}, \, \hat{R} \rangle |,
		\end{aligned}
	\end{equation*}
	where we used the triangular inequality and Lemma \ref{lem:1}.

	$| \langle \bar{\mu}, \, \bar{R} \rangle - \langle \hat{\mu}, \, \hat{R} \rangle |$
	can be further bounded by 
	\begin{equation*}
		\begin{aligned}
		&& |\langle \bar{\mu}, \bar{R} \rangle - \langle \hat{\mu}, \hat{R} \rangle | 
			& = \left|  \max_{\| R\| \le 1}{\langle \bar{\mu}, R \rangle} 
			- \max_{\| R\| \le 1}{\langle \hat{\mu}, R \rangle} \right| \\
		&&  & \le  \max_{\| R\| \le 1}{\langle \bar{\mu} - \hat{\mu}, R \rangle}  \\
		&&  & \le  (R_{max}-R_{min}) \sum_{x \in \mathcal{X}} |\mu(x) - \hat{\mu}(x)|  \\
		&&  & =   2(R_{max}-R_{min}) d_{var}(\bar{\mu}, \hat{\mu}),
		\end{aligned}
	\end{equation*}
	where we used the definition of optimal and estimated values, 
	Lemma \ref{lem:3} for a dual norm, the definition of an inner product, 
	and the definition of $d_{var}(\cdot)$. 
\end{proof}
\begin{proof} (Theorem \ref{thm:main2})
	Combining the aforementioned two inequalities and Lemma \ref{lem:2},
	we get
	\begin{equation*}
		\begin{aligned}
		&& |\langle \bar{\mu}, \hat{R} \rangle - \langle \bar{\mu}, \bar{R} \rangle| 
		 &\le 3(R_{max}-R_{min})d_{var}(\bar{\mu}, \, \hat{\mu}) \\
		&&  &\le 3\sqrt{2} (R_{max}-R_{min}) \sqrt{ O\left( \frac{1}{n} \right) 
			+ O\left( \frac{nd}{N}(1+\frac{1}{\sqrt{\delta}}) \right) }.
		\end{aligned}
	\end{equation*}
\end{proof}

\end{document}